\documentclass[conference]{IEEEtran}
\IEEEoverridecommandlockouts
\usepackage{hyperref}
\usepackage{cite}
\usepackage{amsmath,amssymb,amsfonts}
\usepackage{algorithm}
\usepackage{subcaption}
\usepackage{graphicx}
\usepackage{siunitx}
\usepackage{textcomp}
\usepackage{xcolor}
\usepackage[noend]{algpseudocode}
\usepackage[bottom]{footmisc}
\def\BibTeX{{\rm B\kern-.05em{\sc i\kern-.025em b}\kern-.08em
    T\kern-.1667em\lower.7ex\hbox{E}\kern-.125emX}}
\begin{document}

\title{ NightOwl:\\ Robotic Platform for Wheeled Service Robot\\
}

\author{\IEEEauthorblockN{ Resha Dwika Hefni Al-Fahsi\textsuperscript{1}, Kevin Aldian Winanta\textsuperscript{2}, Fauzan Pradana\textsuperscript{3}, Igi Ardiyanto\textsuperscript{4}, Adha Imam Cahyadi\textsuperscript{5}
}
	
\IEEEauthorblockA{
\textit{Department of Electrical and Information Engineering} \\
\textit{Universitas Gadjah Mada}\\
Yogyakarta, Indonesia \\
\{\textsuperscript{1}resha.dwika.hefni.alfahsi, \textsuperscript{2}kevin.aldian.winanta, \textsuperscript{3}fauzan.p\}@mail.ugm.ac.id, \{\textsuperscript{4}igi, \textsuperscript{5}adha.imam\}@ugm.ac.id
}
\\
}
\maketitle


\begin{abstract}
NightOwl is a robotic platform designed exclusively for a wheeled service robot. The robot navigates autonomously in omnidirectional fashion movement and equipped with LIDAR to sense the surrounding area. The platform itself was built using the Robot Operating System (ROS) and written in two different programming languages (C++ and Python). NightOwl is composed of several modular programs, namely hardware controller, light detection and ranging (LIDAR), simultaneous localization and mapping (SLAM), world model, path planning, robot control, communication, and behaviour. The programs run in parallel and communicate reciprocally to share various information. This paper explains the role of modular programs in the term of input, process, and output. In addition, NightOwl provides simulation visualized in both Gazebo and RViz. The robot in its environment is visualized by Gazebo. Sensor data from LIDAR and results from SLAM will be visualized by RViz.
\\ 
\end{abstract}

\begin{IEEEkeywords}
Service Robot, ROS, Wheeled Mobile Robot, Robotic Platform
\end{IEEEkeywords}

\section{Introduction}
Public places can be defined as places that can be visited by anyone for certain needs and pleasure \cite{b1}. Every day, people do activities such as work, exercise, go to school, and take a stroll to achieve their social and biological needs. People can use the facilities available in their homes or certain public places when carrying out their daily activities. These facilities are used as a means of meeting the needs craved by humans. In public areas, the facilities are spread over at several points. To provide clear information about the location where the facility is located there is a floor plan at the public place. In addition to the floor plan, there is a signpost so that the visitors have knowledge of the path that must be taken to a certain location.

Information such as floor plan, history, and agenda of activities in regard to the public place is very necessary for visitors. Such information can be obtained through signposts scattered in public places. In addition to the signpost, detailed information was obtained through the information center. The information center has a good impact on visitors and managers of public places such as adding to the income of the manager because many visitors are driven to a unique facility in the place so they spend money on retribution or souvenirs, visitor satisfaction on their visit, and visitor curiosity about other facilities besides their main destination\cite{b2}.

One of the problems found in public places such as airport, terminal, hotel, and the office is the visitor's curiosity about the layout and history of these places \cite{b3}. Information about the layout and history of public places can reveal the interesting things hidden in the place. As mentioned before, interesting things about a public place can encourage many visitors to come to that place. With a large number of visitors, the manager of a public place is responsible for solving various problems that are complained by visitors, especially information about the location and the path that must be taken to get to the facilities in that public place. Although visitors know the location of facilities in a certain public place, they will be confused and lost, if the visitors do not know the path that must be taken. 

The problem of visitors' need in the necessary information in regard to the public places, in general, has been resolved by the management of public places with the installation of information signs and road signs. But sometimes the signs are confusing \cite{b4}, giving too little information or too much information on a sign \cite{b5} so that visitors can still be confused or lost. Another solution for information signposts in public places is to make an attractive and interactive digital signpost \cite{b6} so that very much information can fit in a signpost. The making of the digital signpost is quite solutive for the "getting information for the visitors" problem such as the location of a facility but the signpost is static so it cannot solve the "guiding the visitors to certain location" problem. The management of public places in addition to using the signpost, provided an information center so that visitors can ask in detail about the desired information. However, the information center located in public places is less than the maximum in providing services because of limited human resources in serving large numbers of visitors. Information center staff can experience loss of focus due to fatigue in serving the needs of these visitors for a long time continuously \cite{b7}.

On the other hand, there is a service robot. A service robot is robot that is useful for performing certain services that make human work easier. Service robots have a variety of forms and tasks. If seen from its shape, the service robot can be humanoid\cite{b8}, humanoid wheeled\cite{b9}, wheeled\cite{b10} or another decent forms. In carrying out the duties, the service robot carries out repetitive activities that are quite tedious and tiring for humans. These activities can take the form of housework, restaurant or hotel waiter, receptionist and care for people who need special care such as the elderly, people who are sick and disabled \cite{b11}. In these activities sometimes the robot must have the ability to interact with humans. In robotics, there is the term Human-Robot Interaction (HRI) which can be defined as the study of the reciprocal relationship between robot and human to achieve a certain goal. The robot equipped with this feature can make an interaction that can help humans to get information, services, or entertainment. 

In carrying out the duties, the service robot never complains to work for 24 hours. The robot also doesn't need to eat and sleep. So, when compared to human for working every day, it will be cheaper and more humane because there are no labor rights violated, food costs, salaries, family, and other daily human needs. The robot has two main tasks, namely delivering and interacting with the visitors in a building so that the problem which the robot will face is the ability to navigate and interact with humans. Therefore, a service robot can be seen as a solution to overcome the "getting information for the visitors" and "guiding the visitors to certain location" problems at once.

The rest of this paper is organized as follows. In section \ref{RHD}, the robot hardware design is briefly explained. In section \ref{MP}, the modular programs that construct the robotic platform are elaborated. Section \ref{SE} reports the simulation and the result of the robotic platform performance in various specific tasks. In the end, the conclusions and future works are stated in section \ref{C}.

\section{Robot Hardware Design}
\label{RHD}
In their paper\cite{b8}, Hashimoto et al. developed humanoid receptionist service robot. The robot resembles an adult woman who served as a receptionist. The robot named SAYA is able to have simple conversations like human who can show facial expression with certain emotions so that the conversation becomes more interactive. However, the robot can only sit in the information center. The robot cannot move its hands, walk or even stand.

In their paper\cite{b9}, Zhaohui et al. from Xi'an Jiaotong University (XJTU) developed humanoid wheeled service robot. The robot is able to do simple interactions through voice commands. In addition, with the somatosensory system the robot is able to recognize commands through body gesture. The robot can also navigate in the hallway of the building to escort the visitors to certain place and make interactive hand movements.

In their paper\cite{b10}, Khantharak et al. developed wheeled service robot. The robot, named Blackbot, is placed in the lobby of the university building to serve visitors who need help. The robot is able to navigate in the university building so that visitors can be guided to the desired location. To be able to interact with visitors, the robot is equipped with a touch screen that contains a simple conversation about university information and provide responses in the form of sound. The touch screen on the robot shows a funny face in its idle state which make the robot looks friendly.

NightOwl was developed on the wheeled service robot. The wheeled service robot form was chosen because its mobility is well designed for flat terrain\cite{b12}. The wheeled service robot has some advantages in an effort to solve the "guiding the visitors to certain location" problem such as its fast moving speed and its kinematic control way more simple than the other form. These advantages are the reason of the most popular locomotion amongst another possible robot locomotion is the wheeled one.

\subsection{Robot Body Design}
\begin{figure}[htbp]
	\centerline{\includegraphics[width=30mm,scale=0.5]{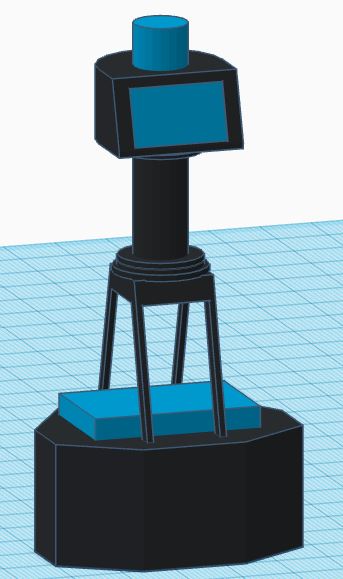}}
	\caption{Robot Body Design.}
	\label{robot_hardware}
\end{figure}

Robot body is a robot part that support other robot parts, such as actuators, sensors, and processors. The robot body chosen must have a strong structure. Therefore, the chosen robot body made of strong and lightweight aluminum. The part that makes this robot tall is the cylindrical support for other parts such as LIDAR which need a certain height to work optimally. This support is made of aluminum, acrylic, and PVC. Figure \ref{robot_hardware} shows the robot body design.

\begin{figure}[htbp]
	\centerline{\includegraphics[width=45mm,scale=0.5]{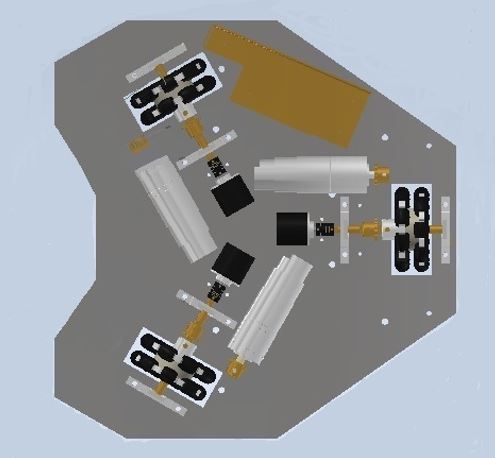}}
	\caption{Robot Base Design.}
	\label{base}
\end{figure}

This robot uses three \ang{90} Swedish wheels with a diameter of 10 cm and eighteen rollers. All three wheels are connected to the motor using chain mechanism. The rotary encoder is connected to the motor through shaft. Wheels, motors, rotary encoders, and battery are also placed on the robot base whose configuration details are shown in figure \ref{base}.

\subsection{Electronic System}
\begin{figure}[htbp]
	\centerline{\includegraphics[height=120mm,width=90mm,scale=3.0]{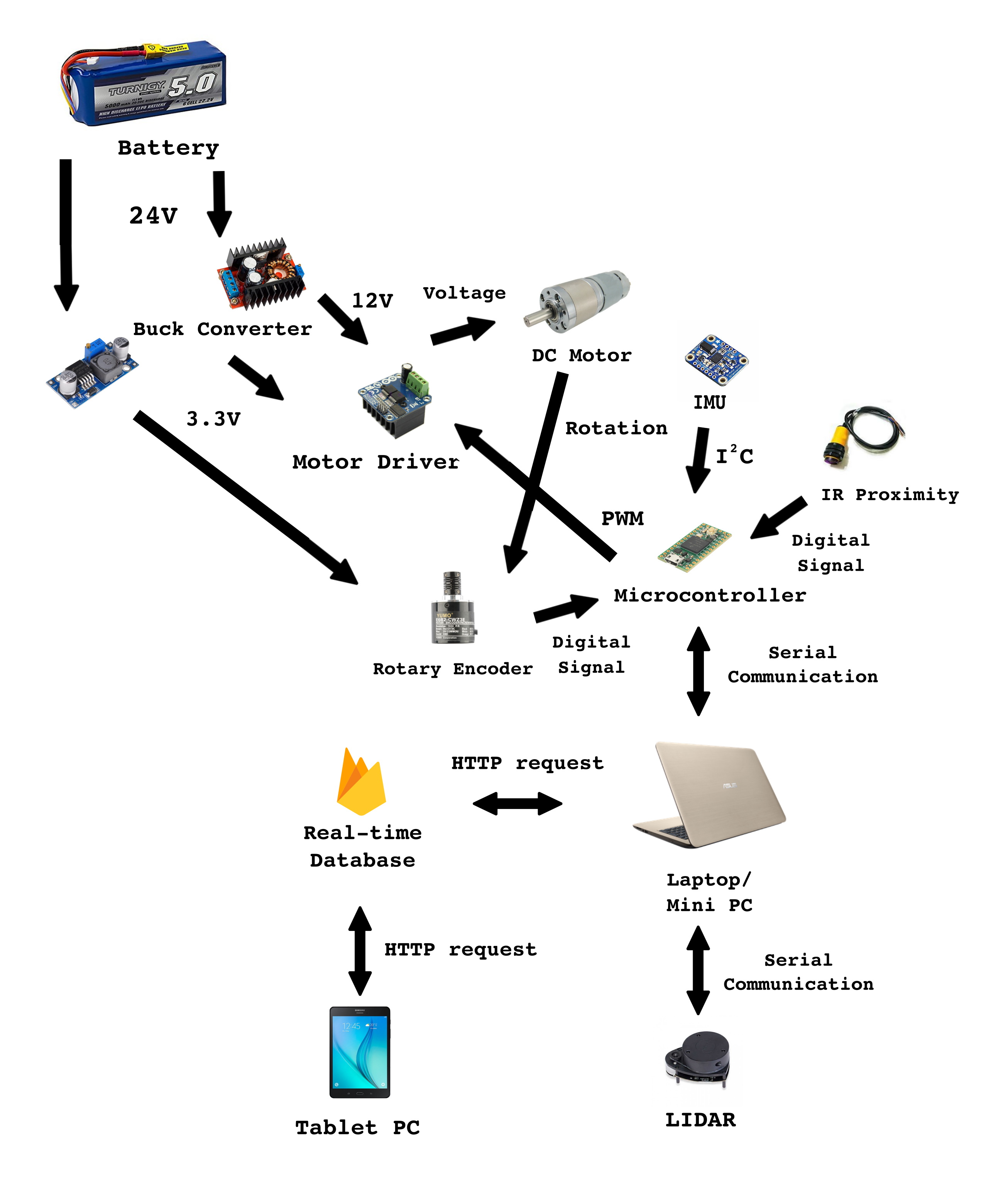}}
	\caption{Electronic System Design.}
	\label{elcsys}
\end{figure}

The electronic system consists of actuators, sensors, controllers, networking and power supply. The actuators move the robot by applying electrical signals to generate some force. The actuators consist of DC motors. The sensors perceived the environment by transforming physical phenomena into electrical signals. The sensors consist of inertial measurement unit (IMU), LIDAR, rotary encoders and infrared (IR) proximity sensor. The controllers consist of microcontroller and main controller. The microcontroller receives the electrical signal from the sensors and emits electrical signals to the actuators through motor drivers. The main controller consists of a tablet PC and a laptop or mini PC. 
The networking system provides a real-time database that stores the amount of crucial information such as requested room by Human-Robot Interaction (HRI) and robot pose by laptop or mini PC.
The power supply provides electrical power for another electrical component. The power supply consists of battery and buck converters. The buck converters supply lower voltage from the battery so that the electronic components work in the appropriate specification. The electronic system design is shown in figure \ref{elcsys}.

\section{Modular Program}
\label{MP}
\begin{figure}[htbp]
	\leftline{\includegraphics[width=100mm,scale=1.0]{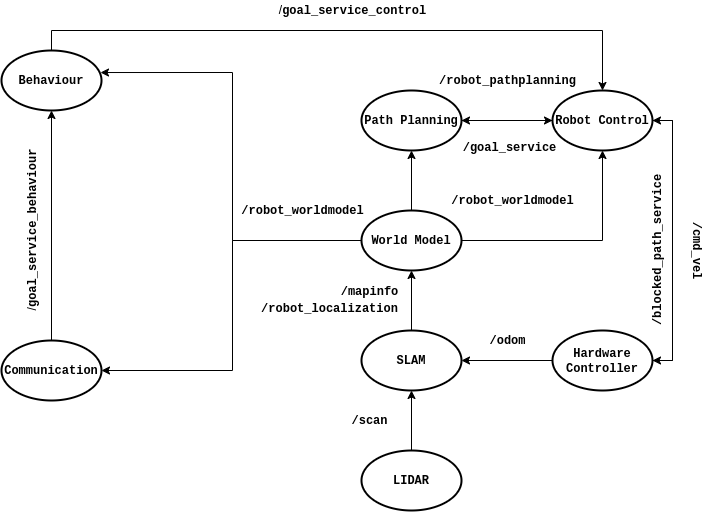}}
	\caption{Modular Program Architecture.}
	\label{diagram}
\end{figure}

This ROS based platform architecture has an advantage in its modularity so that the algorithms can run in parallel. In the term of ROS, this modular program also called {\fontsize{9}{1.2}{\fontfamily{qcr}{\selectfont {node}}}}. The {\fontsize{9}{1.2}{\fontfamily{qcr}{\selectfont {node}}}} can exchange information via {\fontsize{9}{1.2}{\fontfamily{qcr}{\selectfont {topic}}}}. The {\fontsize{9}{1.2}{\fontfamily{qcr}{\selectfont {topic}}}} provides information in continuous way which called {\fontsize{9}{1.2}{\fontfamily{qcr}{\selectfont {message}}}}. The topic also provides information in a one-time called fashion which called {\fontsize{9}{1.2}{\fontfamily{qcr}{\selectfont {service}}}}. The {\fontsize{9}{1.2}{\fontfamily{qcr}{\selectfont {node}}}} can obtain information from the {\fontsize{9}{1.2}{\fontfamily{qcr}{\selectfont {topic}}}} by declaring the subscriber and transmits information by declaring the publisher. The modular program architecture is shown in figure \ref{diagram}.

\subsection{Hardware Controller}
\label{HW}
This {\fontsize{9}{1.2}{\fontfamily{qcr}{\selectfont {node}}}} handles reciprocal Universal Asynchronous Receiver-Transmitter (UART) serial communication between the microcontroller and the laptop or mini PC. This serial communication has 115200 baud which means it has the capability to transfer 115200 symbols per second. This high enough amount of baud rate was chosen to keep the data transfer in real-time as possible while maintaining the reliability of the data \cite{b13}. Although serial communication is asynchronous, the serial data is received in the synchronous patterns of 8 bit symbol by encapsulating the data within a consecutive serial data sequence which is consists of header, body and tail. The header indicates the serial data sequence is initiated, the body is the crux of the serial data sequence and the tail tells the serial data sequence is over.

As mentioned before, the microcontroller handles the process in regard to the actuators and the sensors. The laptop or mini PC sends some requests corresponding to manipulate the movement of the actuator and receives a bunch of processed sensor data in the form of odometry data. The sensors data are processed by the microcontroller obtained from the IMU the rotary encoders and IR proximity sensor. The IR proximity sensor triggered whenever the infrared beam hit an object. In the implementation, this event raises a blocked flag through {\fontsize{9}{1.2}{\fontfamily{qcr}{\selectfont {/blocked\_path\_service}}}} {\fontsize{9}{1.2}{\fontfamily{qcr}{\selectfont {topic}}}}. The IMU is an electronic component that combines accelerometer, gyroscope and magnetometer to get the orientation of an object. The rotary encoders data are in the form of the amount of instantaneous pulse at time $t$ which are denoted by $\xi$. By using these data, the magnitude of the linear velocity of the wheel $\dot{q}_{n}$ at time $t$ can be calculated as follows.
\begin{align}
\dot{q}_{n}(t) = \frac{\pi \cdot d}{ppr} \cdot \xi
\end{align}

Where $d$ is the diameter of the wheel and $ppr$ is an abbreviation of pulse per revolution. Basically, the rotary encoder is a sensor that changes the rotation movement to form a voltage pulse. This sensor emits a pulse with a certain amount of each spin. With this information, the magnitude of the velocity and position of the rotating object can be known. Then, the rotary encoder has a specification of how many pulse emitted every single spin or revolution. This rate of the amount of pulse every single revolution is called pulse per revolution. The magnitude of the linear velocity of the wheel is required to compute odometry of the robot by using forward kinematics. Then, the odometry data from microcontroller is published through {\fontsize{9}{1.2}{\fontfamily{qcr}{\selectfont {$/$odom}}}} {\fontsize{9}{1.2}{\fontfamily{qcr}{\selectfont {topic}}}}.

The requests from laptop or mini PC are processed and obtained from {\fontsize{9}{1.2}{\fontfamily{qcr}{\selectfont {$/$cmd$\_$vel}}}} {\fontsize{9}{1.2}{\fontfamily{qcr}{\selectfont {topic}}}} which contains the desired velocity for the robot. The desired velocity is going to be translated to the magnitude of the linear velocity of the wheel by using inverse kinematics. 

\begin{figure}[htbp]
	\centerline{\includegraphics[width=50mm,scale=0.5]{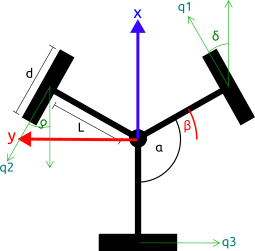}}
	\caption{Top View of the Three \ang{90} Swedish Wheels Configuration.}
	\label{conf}
\end{figure}

The robot with a three \ang{90} Swedish wheels configuration shown in figure \ref{conf} can move in all directions directly without changing its orientation. This can happen because there are rollers on the \ang{90} Swedish wheel. These rollers make the wheel can shift in all directions.

There are several parameters for the robot with this configuration, namely $d$ which is the diameter of the wheels, $L$ which is the distance from the wheels to the center of the robot, $\delta$ which is the angle between the positive direction of the rightmost wheel linear velocity vector and the positive direction of the $x$ component of the whole robot velocity vector which the direction of the wheel linear velocity vector is positive when rotating in clockwise which observed from the outside of the robot and $\alpha$ which is the angle between the axle shaft.

In order to manipulate the robot movement with this configuration, there are two kinds of mathematical process regarding the kinematics of the actuators namely the inverse kinematics and forward kinematics. With the inverse kinematics, the linear velocity of each wheel is calculated from the velocity of the whole robot, whereas with the forward kinematics, the velocity of the whole robot is calculated from the magnitude of the linear velocity of each wheel. The inverse kinematics equation for this configuration is formulated as follows.
\begin{align}
\label{equation}
\dot{q}(t) = B \cdot V_{R}(t)
\end{align}

Whereas the forward kinematics equation can be formulated as follows.
\begin{align}
V_{R}(t) = B^{-1} \cdot \dot{q}(t)
\end{align}

Where, $\dot{q}(t)$ is linear velocity of the wheel at time $t$ and $V_{R}(t)$ is velocity of the whole robot at time $t$ in local coordinate. The matrix form of $\dot{q}(t)$, $V_{R}(t)$ and $B$ can be defined as follows.
\begin{align}
\dot{q}(t) = \begin{bmatrix}
\dot{q}_{1}(t)\\ 
\dot{q}_{2}(t)\\
\dot{q}_{3}(t)\\ 
\end{bmatrix}
\end{align}

\begin{align}
V_{R}(t) = \begin{bmatrix}
R_{\dot{x}}(t)\\ 
R_{\dot{y}}(t)\\
R_{\dot{\theta}}(t)\\ 
\end{bmatrix}
\end{align}

\begin{align}
B = \begin{bmatrix}
\label{matrix_b}
\cos{(\delta)} & \sin{(\delta)} & L\\
-\cos{(\delta)} &  \sin{(\delta)} & L\\
0 & -1 & L\\
\end{bmatrix}
\end{align}

The matrix $B$ is the inverse kinematics transformation matrix which derived from the following matrix.
\begin{align}
\begin{bmatrix}
\cos{(\vartheta_{1})} & \sin{(\vartheta_{1})} & L\\
\cos{(\vartheta_{2})} & \sin{(\vartheta_{2})} & L\\
\cos{(\vartheta_{3})} & \sin{(\vartheta_{3})} & L\\
\end{bmatrix}
\end{align}

This matrix is derived from the kinematics relationship between the whole robot velocity vector and each of the wheels linear velocity vector \cite{b14}. By applying equation \ref{equation}, each row of the matrix can be expressed as follows.
\begin{align}
\dot{q}_{n}(t) = \cos{(\vartheta_{n})} \cdot R_{\dot{x}}(t) + \sin{(\vartheta_{n})} \cdot R_{\dot{y}}(t) + L \cdot R_{\dot{\theta}}(t)
\end{align}

\begin{figure}[htbp]
	
	\begin{subfigure}{0.5\textwidth}
		\centerline{\includegraphics[width=0.3\linewidth, height=3cm]{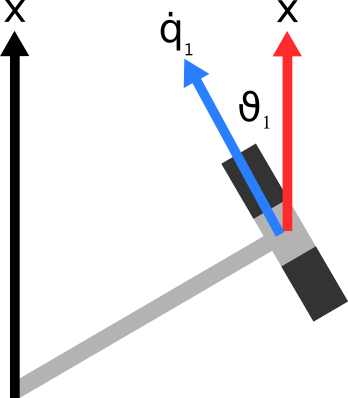}}
		\caption{Rightmost Wheel Configuration.}
		\label{wheel1}
	\end{subfigure}
	\begin{subfigure}{0.5\textwidth}
		\centerline{\includegraphics[width=0.3\linewidth, height=3cm]{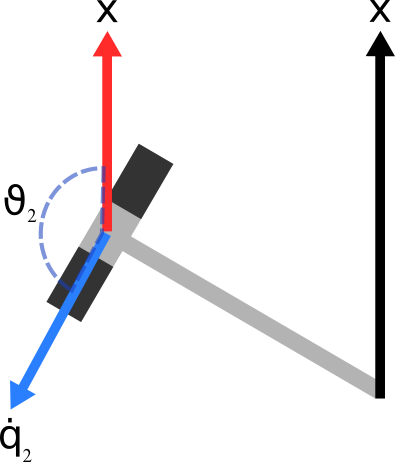}}
		\caption{Leftmost Wheel Configuration.}
		\label{wheel2}
	\end{subfigure}
	\begin{subfigure}{0.5\textwidth}
		\centerline{\includegraphics[width=0.25\linewidth, height=4cm]{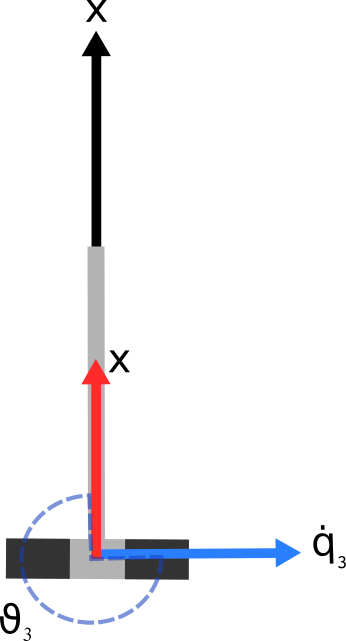}}
		\caption{Rear Wheel Configuration.}
		\label{wheel3}
	\end{subfigure}
	
	\caption{Each of the Possible Wheel Configuration.}
	\label{wheel}
\end{figure}

The equation shows the impact of the linear and angular velocity of the whole robot to the magnitude of the linear velocity of the wheel. The $\cos{(\vartheta_{n})} \cdot R_{\dot{x}}(t) + \sin{(\vartheta_{n})} \cdot R_{\dot{y}}(t) $ part shows the dot product between the two-dimensional unit vector of the linear velocity of the wheel, $[\cos{(\vartheta_{n})}, \sin{(\vartheta_{n})}]^T$, and the two-dimensional vector of the robot linear velocity, $[ R_{\dot{x}}(t), R_{\dot{y}}(t)]^T$. Since the wheel can only spin clockwise and counterclockwise, the unit vector is crucial for defining the linear velocity of the wheel would go and its magnitude. The dot product maps the magnitude of the robot linear velocity vector in the perspective of the unit vector of the linear velocity of the wheel. Thus, this dot product yields the magnitude of the linear velocity of the wheel. Each of the two-dimensional unit vectors of the wheels is represented in the term of cosine and sine with the direction of $\vartheta_{n}$ relative to the robot reference frame
which is denoted by $\vartheta_{1}$, $\vartheta_{2}$, and $\vartheta_{3}$ respectively. Considering equation \ref{matrix_b}, both of the variables $\vartheta_{1}$ and $\delta$ have the same value, which is \ang{30}. Henceforth, the value of $\vartheta_{2}$ and $\vartheta_{3}$ are equal to $\vartheta_{1} + \ang{120}$ and $\vartheta_{1} + \ang{240}$ respectively. The depiction regarding the wheel configuration is shown in figure \ref{wheel}

The angular velocity of the whole robot plays an important part in increasing the magnitude of the linear velocity of the wheel too. The $L \cdot R_{\dot{\theta}}(t)$ part is the tangential velocity perpendicular to the angular velocity of the robot. Thus, any small value in the robot angular velocity affects the magnitude of the linear velocity of the wheel. Therefore, in order to reduce the misreading due to uncertainty of rotary encoders, the heading data from IMU can be used either as a proxy of the $R_{\dot{\theta}}(t)$ or by making use of certain sensor fusion algorithm along with the encoders data.

As known before, $V_{R}(t)$ is the local coordinate whole robot velocity. In order to get the global coordinate representation, the velocity is calculated as follows.
\begin{align}
\begin{bmatrix}
v_{{x}}(t)\\ 
v_{{y}}(t)\\
\end{bmatrix}
=
\begin{bmatrix}
cos{(\theta_{t-1})} & -sin{(\theta_{t-1})}\\ 
sin{(\theta_{t-1})} & cos{(\theta_{t-1})}\\
\end{bmatrix}
\cdot
\begin{bmatrix}
R_{\dot{x}}(t)\\ 
R_{\dot{y}}(t)\\
\end{bmatrix}
\end{align}

Where the angular velocity could be defined as follows.
\begin{align}
\omega(t) = R_{\dot{\theta}}(t)
\end{align}

By knowing the whole robot velocity in global coordinate, $v_x(t), v_y(t)$ and $\omega(t)$, the odometry of the robot can be defined as the total summation of the velocity from the initial point until the point at time $t$ and it can be computed as follows.
\begin{align}
\begin{bmatrix}
{{x}}(t)\\ 
{{y}}(t)\\
{{\theta}}(t)\\ 
\end{bmatrix}
= 
\begin{bmatrix}
{{x}}(t-1)\\ 
{{y}}(t-1)\\
{{\theta}}(t-1)\\ 
\end{bmatrix}
+
\begin{bmatrix}
v_{{x}}(t)\\ 
v_{{y}}(t)\\
{{\omega}}(t)\\ 
\end{bmatrix}
\end{align}

\subsection{LIDAR}
\begin{figure}[htbp]
	\centerline{\includegraphics[width=60mm,scale=1.0]{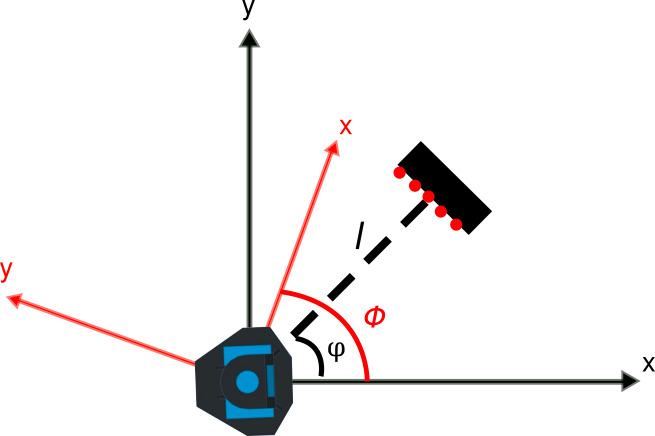}}
	\caption{LIDAR Measurement Representation.}
	\label{lidar}
\end{figure}
LIDAR is a sensor used to measure distances. LIDAR emits light then measures the time the reflection of the light to determine the distance of the object. This {\fontsize{9}{1.2}{\fontfamily{qcr}{\selectfont {node}}}} process reading from LIDAR sensor into data represented by a set $z_{t}$ consist of $N$ pair of the range and angle. The data are sent to another {\fontsize{9}{1.2}{\fontfamily{qcr}{\selectfont {node}}}} through  {\fontsize{9}{1.2}{\fontfamily{qcr}{\selectfont {$/$scan}}}}  {\fontsize{9}{1.2}{\fontfamily{qcr}{\selectfont {topic}}}}. As shown in the figure \ref{lidar}, the range data can be interpreted as Euclidean distance of a point and always come in pair with the angle data which indicates the direction of the point in the perspective of local coordinate in two-dimensional space. Then, to calculate a point $(x_t, y_t)$ using the range $l_t^{[n]}$ and angle $\varphi_t^{[n]}$ data can be formulated as follows.
\begin{align}
\label{xlidar}
x_t = l_t^{[n]} \cdot \cos{(\varphi_t^{[n]})}\\
\label{ylidar}
y_t = l_t^{[n]} \cdot \sin{(\varphi_t^{[n]})}
\end{align}
 
In some cases, the LIDAR needs to be adjusted by rotating it due to some physical constraints. The calculated point also has to be rotated in order to get valid data. The LIDAR can be rotated with arbitrary angle $\phi$. Then, the point can be formulated as follows.

\begin{align}
\label{rotatedlidar}
\begin{bmatrix}
x_t^{rotated}\\
y_t^{rotated}
\end{bmatrix}
=
\begin{bmatrix}
\cos{(\phi)} & -\sin{(\phi)}\\
\sin{(\phi)} & \cos{(\phi)}
\end{bmatrix}
\cdot
\begin{bmatrix}
x_t\\
y_t
\end{bmatrix}
\end{align}   

\vspace{10pt}

\subsection{SLAM}
\label{SLAM}
This {\fontsize{9}{1.2}{\fontfamily{qcr}{\selectfont {node}}}} process published odometry data from {\fontsize{9}{1.2}{\fontfamily{qcr}{\selectfont {$/$odom}}}} {\fontsize{9}{1.2}{\fontfamily{qcr}{\selectfont {topic}}}} and LIDAR data from {\fontsize{9}{1.2}{\fontfamily{qcr}{\selectfont {$/$scan}}}} {\fontsize{9}{1.2}{\fontfamily{qcr}{\selectfont {topic}}}}. By combining these data, the robot pose and its perceived environment could be computed with SLAM. SLAM is famous in regard to its chicken and egg problem in robotics that is trying to solve. The problem itself is how to create a map of the environment and determine the robot position in the map concurrently. The map could be constructed with the occupancy grid mapping algorithm and the robot pose could be calculated with the infamous particle-based localization, Monte Carlo Localization (MCL). SLAM with MCL shows better performance than the most common approach using Extended Kalman Filter (EKF) \cite{b15}. The constructed map then published via {\fontsize{9}{1.2}{\fontfamily{qcr}{\selectfont {$/$mapinfo}}}} {\fontsize{9}{1.2}{\fontfamily{qcr}{\selectfont {topic}}}} and the robot pose is published through {\fontsize{9}{1.2}{\fontfamily{qcr}{\selectfont {$/$robot$\_$localization}}}} {\fontsize{9}{1.2}{\fontfamily{qcr}{\selectfont {topic}}}}.

As mentioned before, Monte Carlo Localization is a localization method based on the nonparametric implementation of the Bayes filter so that it represents the belief, internal knowledge about the state of the environment, by a set of particles \cite{b16}. The set of particles $\chi_{t}$ consists of $M$ particles $\{p_{t}^{[1]}, p_{t}^{[2]}, p_{t}^{[3]},..., p_{t}^{[M]} \}$. Each of the particles $p_{t}^{[m]}$ has three states, $x_t$, $y_t$ and $\theta_t$. These state are updated over time by $motion\_ model$ algorithm which is described the uncertainty of movement obtained from odometry data which is consists of velocity of the robot in global coordinate, $v_x(t)$, $v_y(t)$ and $\omega(t)$. The $motion\_ model$ algorithm of the robot is shown in algorithm \ref{motionmodel}.

\begin{algorithm}
\caption{Robot Motion Model}
\begin{algorithmic}[1]
	\Function{Motion\_Model}{$v_x(t), v_y(t), \omega(t), \chi_{t-1}$}
	
	\State $
			{\chi}_{t} \leftarrow \{\}$
	
	\For{$m \leftarrow 1$ to $M$}
	
	\State $d_x$ $\leftarrow$ $v_x(t)$ $\cdot$ $\cos{(\theta_{p_{t-1}^{[m]}})}$ $+$ $v_y(t)$ $\cdot$ $\sin{(\theta_{p_{t-1}^{[m]}})}$
	\State $d_y$ $\leftarrow$ $-$ $v_x(t)$ $\cdot$ $\sin{(\theta_{p_{t-1}^{[m]}})}$ $+$ $v_y(t)$ $\cdot$ $\cos{(\theta_{p_{t-1}^{[m]}})}$
	
	\State $t_x$ $\leftarrow$ 
	 $d_x$ $\cdot$ $ $
	 $r \sim
	 \mathcal{N}(1.0 - \bar{\sigma}_{x}, \sigma_{x})$ 
	\State $t_y$ $\leftarrow$ 
	$d_y$ $\cdot$ $ $
	$r \sim
	 \mathcal{N}(1.0 - \bar{\sigma}_{y}, \sigma_{y})$ 
	
	\State $d_x$ $\leftarrow$ $t_x$ $\cdot$ $\cos{(\theta_{p_{t-1}^{[m]}})}$ $-$ $t_y$ $\cdot$ $\sin{(\theta_{p_{t-1}^{[m]}})}$
	\State $d_y$ $\leftarrow$ $t_x$ $\cdot$ $\sin{(\theta_{p_{t-1}^{[m]}})}$ $+$ $t_y$ $\cdot$ $\cos{(\theta_{p_{t-1}^{[m]}})}$
	
	\State $x_{p_t^{[m]}} \leftarrow x_{p_{t-1}^{[m]}} + d_x$
	\State $y_{p_t^{[m]}} \leftarrow y_{p_{t-1}^{[m]}} + d_y$ 
	\State $\theta_{p_t^{[m]}} \leftarrow \theta_{p_{t-1}^{[m]}} + \omega(t)$ $\cdot$ $ $
	$r \sim \mathcal{N}(1.0 - \bar{\sigma}_{\theta}, \sigma_{\theta})
	$ 
	
	\State normalize $\theta_{p_t^{[m]}}$ into the range of $\ang{0}$ to $\ang{360}$
	
	\State $
			{\chi}_t \cup p_t^{[m]}$ 
	
	\EndFor
	\State \textbf{end for}
	
	\State \Return $
					{\chi}_{t}$

	\EndFunction
	\State \textbf{end function}
\end{algorithmic}
\label{motionmodel}
\end{algorithm}

The notion behind the $motion\_model$ algorithm is modeling the uncertainty from the encoders and the IMU. Then, from both of the sensors, the velocity data is obtained. The velocity data is transformed from global coordinate form into local coordinate form and multiplied by the noise factor $r$ drawn from a normal distribution. The distribution is centered at $1.0 - \bar{\sigma}$ by an appropriate standard derivation $\sigma$ with respect to each of the velocity components. The parameter $\bar{\sigma}$ exhibits the inherent uncertainty characteristic of the average velocity of the robot whilst the parameter $\sigma$ is the uncertainty of the velocity that might be reached.  
Then, the velocity is transformed back to the global coordinate representation in order to update the old state of the particle.

The LIDAR data is processed in occupancy grid mapping to construct map $\mathcal{M}$ and in the $measurement\_model$ algorithm to weight the particles. The particles are weighted based on the range from LIDAR $l_t^{[n]}$ and the relative range from each of the particles $l_{p_t^{[m]}}^{[n]}$. There is a parameter $\varepsilon$ which is defined as the difference between the range from LIDAR and the relative range from each of the particles at $\varphi_t^{[n]}$ direction. Then, $\varepsilon$ multiplied by $\frac{1}{p_{hit}}$. The $p_{hit}$ is defined as probability of each of the particles using the range $l_t^{[n]}$ at $\varphi_t^{[n]}$ direction hit the obstacle. The result of multiplication of $\epsilon$ and $\frac{1}{p_{hit}}$ is plugged into the following formula.
\begin{align}
\label{fx}
f(x) = x^8 + e^x
\end{align}

\begin{figure}[htbp]
	\centerline{\includegraphics[width=90mm,scale=1.0]{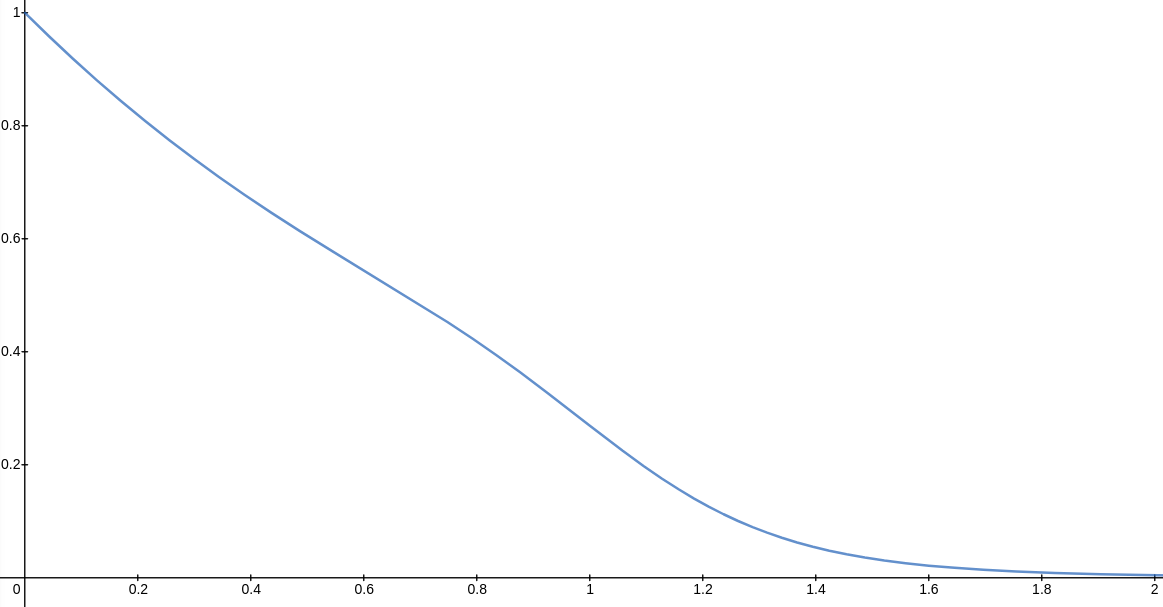}}
	\caption{Particle Weight Distribution.}
	\label{weight}
\end{figure}

Then, the outcome of the formula is added by all of the outcomes of the formula with another calculated parameter from the LIDAR data and divided by $N$ to get the mean of the processed LIDAR data $\bar{z}$. The final weight of each of the particles is defined by $\frac{1}{\bar{z}}$ 
and normalized so that $\sum_{m=1}^{M} weight_{p_t^{[m]}} = 1$. 
The distribution of the weight is shown in figure \ref{weight}.

\begin{algorithm}
	\caption{Robot Measurement Model}
	\begin{algorithmic}[1]
		
		\Function{Measurement\_Model}{$z_t, \chi_{t}, \mathcal{M}$}
		
		\State $\overline{\chi}_{t} \leftarrow \{\}$
		\State $total\_weight\leftarrow 0$
		
		\For{$m \leftarrow 1$ to $M$}
		\State $\bar{z} \leftarrow 0$
		\For{$n \leftarrow 1$ to $N$}
		\State calculate $x_t$ and $y_t$ using eq. \ref{xlidar} and \ref{ylidar}
		\State calculate $x_t^{rotated}$ and $y_t^{rotated}$ using eq. \ref{rotatedlidar}
		\State transform $x_t^{rotated}$ and $y_t^{rotated}$ from local coordinate to global coordinate
		\State $p_x \leftarrow x_{p_t^{[m]}} + x_t^{rotated}$
		\State $p_y \leftarrow y_{p_t^{[m]}} + y_t^{rotated}$   
		\State calculate $p_{hit}$ based on $p_x$ and $p_y$ on $\mathcal{M}$
		\State $l_{p_t^{[m]}}^{[n]} \leftarrow 0$
		\While{$l_{p_t^{[m]}}^{[n]} \leq l_{max}$ }
		\State calculate $x_t$ and $y_t$ using eq. \ref{xlidar} and \ref{ylidar} based on $l_{p_t^{[m]}}^{[n]}$
		\State calculate $x_t^{rotated}$ and $y_t^{rotated}$ using eq. \ref{rotatedlidar} 
		\State $p_x \leftarrow x_{p_t^{[m]}} + x_t^{rotated}$
		\State $p_y \leftarrow y_{p_t^{[m]}} + y_t^{rotated}$
		\If{$HitObstacle(p_x, p_y, \mathcal{M})$}
		\State \textbf{break}
		\EndIf
		\State \textbf{end if}
		\State $l_{p_t^{[m]}}^{[n]} \leftarrow l_{p_t^{[m]}}^{[n]} + dl $ \Comment increment with a small change $dl$
		\EndWhile
		\State \textbf{end while}
		\If{$l_{p_t^{[m]}}^{[n]} < l_{max}$}
		\State $\epsilon \leftarrow \mid l_t^{[n]} - l_{p_t^{[m]}}^{[n]} \mid$
		\State calculate $result$ by passing $\epsilon \cdot \frac{1}{p_{hit}}$ into the eq. \ref{fx}  
		\State $\bar{z} \leftarrow \bar{z} + result$
		\Else
		\State $\bar{z} \leftarrow \bar{z} + E $ \Comment $E$ is a very big number
		\EndIf
		\State \textbf{end if}
		\EndFor 
		\State \textbf{end for}
		\State $\bar{z} \leftarrow \frac{\bar{z}}{N}$
		\State $weight_{p_t^{[m]}} \leftarrow \frac{1}{\bar{z}}$
		\State $total\_weight\leftarrow total\_weight + weight_{p_t^{[m]}}$
		\EndFor
		\State \textbf{end for}
		
		\State $w_{avg} \leftarrow \frac{total\_weight}{M}$
		\State $w_{slow} \leftarrow w_{slow} + \alpha_{slow} \cdot (w_{avg} - w_{slow})$
		\State $w_{fast} \leftarrow w_{fast} + \alpha_{fast} \cdot (w_{avg} - w_{fast})$
		\For {$m \leftarrow 1$ to $M$}
		\State $weight_{p_t^{[m]}} \leftarrow \frac{weight_{p_t^{[m]}}}{total\_weight}$
		\State $\overline{\chi}_t \cup p_t^{[m]}$ 
		\EndFor
		\State \textbf{end for}
		\State \Return $\overline{\chi}_{t}, w_{fast}, w_{slow}$
		
		\EndFunction
		\State \textbf{end function}
	\end{algorithmic}
	\label{measuremodel}
\end{algorithm}

The $y$-axis of the graph shows the magnitude of the weight and the $x$-axis of the graph shows the result of the multiplication $\varepsilon \cdot \frac{1}{p_{hit}}$. The graph shows the relationship regarding the gap of measured range from the sensor and measured range depends on the particle pose. The greater the gap the smaller the weight. This also emphasized by $\frac{1}{p_{hit}}$ factor so that the smaller the probability of hitting the obstacle the smaller the weight. As from the graph, the weight is ranged from 0 to 1 as same as the appropriate range of the probability scale.

However, LIDAR has some limitations. It has a maximum and minimum reading of the range and angle. The limitation reading of the range is the main concern when calculating the algorithm which involved the LIDAR data. The maximum range reading $l_{max}$ should be ignored. The $measurement\_model$ algorithm with input $z_t = \{(l_t^{[1]}, \varphi_t^{[1]}), (l_t^{[2]}, \varphi_t^{[2]}), (l_t^{[3]}, \varphi_t^{[3]}),...,(l_t^{[N]}, \varphi_t^{[N]}) \}$,  $\chi_{t}$ and $\mathcal{M}$ is shown in 
algorithm \ref{measuremodel}.

From algorithm \ref{measuremodel}, there are new parameters that are introduced. These parameters are required in the resampling process. The resampling process attempts to recover the stray particle back to the expected pose. Adding random particles is one of the solutions to achieve the recovery process. The random particles are added based on the parameters mentioned before. The key parameter of this process is the average weight of the particles $w_{avg}$ which related to the probability of sensor measurements which indicates how well the sensor measures something. The $w_{avg}$ parameter regulate a short-term $w_{fast}$ and a long-term $w_{slow}$ parameters which represent the immediate and delayed value of $w_{avg}$ respectively. The parameters $\alpha_{fast}$ and $\alpha_{slow}$ are discount rates for the immediate and delayed value of $w_{avg}$ respectively which the values are described as $0 < \alpha_{slow} \ll \alpha_{fast}$. Then, the random particles are added with the probability $\max\{\varOmega, 1.0 - \frac{w_{fast}}{w_{slow}} \}$ where the $\varOmega$ is a very small number to keep the resampling process in order to avoid false convergence of the particles. The resampling process is shown in algorithm \ref{resample}.

The random particles spread with the bounded condition. The random pose $x_{rand}$ and $y_{rand}$ are used as a reference in which direction the particle should spread through unit vector normalization. The particle drift factor $\gamma$ tells how far the particle drifts. The random particle orientation is determined by sampling with uniform distribution which modeling the error of the IMU in the range of $\pm \theta_\epsilon$.
The core of the resampling process is shown inline 19 to 25 on the algorithm \ref{resample}. It uses low variance sampling which chooses a random number $r$ that is drawn from a uniform distribution in the range from $0$ to $\frac{1}{M}$ and select those particles at some index $idx$ that correspond to $u = r + (m-1) \cdot \frac{1}{M}$ where $m = 1,...,M$. Thus, this process could be formulated as follows.
\begin{align}
idx = \underset{j}{\mathrm{argmin}} \sum_{m=1}^{j} weight_{p_t^{[m]}} \geq U
\end{align}

Thus, the principle of Monte Carlo Localization algorithm is a sequence of $motion\_model$, $measurement\_model$ and resampling process. Due to its multimodal distribution property, the computed particles need to be clustered and found $k$ centroids, the center-of-mass of a cluster, that exists in the distribution \cite{b17} throughout $n$ epochs. The robot pose $x_t$, $y_t$ and $\theta_t$ are estimated by picking the best amongst the cohort of the centroids $\mathcal{K}$ based on its weight. These centroids are calculated using the Lloyd-type $k$-means clustering algorithm which has $O(Mnk)$ time complexity \cite{b18}. Each of the centroids weighted by computing the average weight of the particles with respect to its cluster \cite{b19}.
The main algorithm of Monte Carlo Localization is shown in algorithm \ref{mcl}

\begin{algorithm}
	\caption{Resampling Process}
	\begin{algorithmic}[1]
		\Function{Resample}{$\chi_{t}, w_{fast}, w_{slow}$}
		\State $\overline{\chi}_{t} \leftarrow \{\}$
		\State $c \leftarrow weight_{p_t^{[1]}}$
		\State $r \sim \mathcal{U}(0.0, \frac{1}{M})$
		\For{$m \leftarrow 1$ to $M$}
		
		\State \textbf{with} probability $\max\{\varOmega, 1.0 - \frac{w_{fast}}{w_{slow}} \}$ \textbf{do}
		\State $ $ $ $ $ $ $ $ $ $ sample random pose $(x_{rand}, y_{rand})$
		\State $ $ $ $ $ $ $ $ $ $ $p_x \leftarrow x_{rand} - x_{p_t^{[m]}}$
		\State $ $ $ $ $ $ $ $ $ $ $p_y \leftarrow y_{rand} - y_{p_t^{[m]}}$
		\State $ $ $ $ $ $ $ $ $ $ $length \leftarrow hypot(p_x, p_y)$\
		\State $ $ $ $ $ $ $ $ $ $ $p_x \leftarrow \frac{p_x}{length}$
		\State $ $ $ $ $ $ $ $ $ $ $p_y \leftarrow \frac{p_y}{length}$
		\State $ $ $ $ $ $ $ $ $ $ $x_{\bar{p}_t} \leftarrow x_{p_t^{[m]}} + \gamma \cdot p_x$
		\State $ $ $ $ $ $ $ $ $ $ $y_{\bar{p}_t} \leftarrow y_{p_t^{[m]}} + \gamma \cdot p_y$
		\State $ $ $ $ $ $ $ $ $ $ $\theta_{\bar{p}_t} \leftarrow \theta_{rand} \sim \mathcal{U}(\theta_{p_t^{[m]}} - \theta_{\varepsilon}, \theta_{p_t^{[m]}} + \theta_{\varepsilon})$
		\State $ $ $ $ $ $ $ $ $ $ $weight_{\bar{p}_t} \leftarrow \frac{1}{M} \cdot weight_{p_t^{[m]}}$
		\State $ $ $ $ $ $ $ $ $ $ $\overline{\chi}_t \cup \bar{p}_t$
		\State \textbf{else}
		\State $ $ $ $ $ $ $ $ $ $ $idx \leftarrow 1$
		\State $ $ $ $ $ $ $ $ $ $ $u \leftarrow r + (m-1) \cdot \frac{1}{M}$
		\State $ $ $ $ $ $ $ $ $ $ $\textbf{while}$ $u > c$ $\textbf{do}$
		\State $ $ $ $ $ $ $ $ $ $ $ $ $ $ $ $ $ $ $ $ $ $ $idx \leftarrow idx + 1$
		\State $ $ $ $ $ $ $ $ $ $ $ $ $ $ $ $ $ $ $ $ $ $ $c \leftarrow c + weight_{p_t^{[idx]}}$
		\State $ $ $ $ $ $ $ $ $ $ \textbf{end while}
		\State $ $ $ $ $ $ $ $ $ $ $\overline{\chi}_t \cup p_t^{[idx]}$ 
		\State \textbf{end with}
		\EndFor
		\State \textbf{end for}
		
		\State \Return $\overline{\chi}_t$ 
		
		\EndFunction
		\State \textbf{end function}
	\end{algorithmic}
	\label{resample}
\end{algorithm}
\vspace{1000pt}
\begin{algorithm}
	\caption{Monte Carlo Localization}
	\begin{algorithmic}[1]
		\Function{MCL}{$v_x(t), v_y(t), \omega(t), z_t, \chi_{t-1}, \mathcal{M}$}
		\State $\chi_{t} \leftarrow motion\_model(v_x(t), v_y(t), \omega(t), \chi_{t-1})$
		\State $\chi_{t}, w_{fast}, w_{slow}$ $\leftarrow$ $ $ $ $ $ $ $ $ $ $ $ $ $ $ $ $ $ $ $ $ $ $ $ $ $ $ $ $ $ $ $ $ $ $ $ $ $ $ $ $ $ $ $ $ $ $ $ $ $ $ $ $ $ $ $ $ $ $ $ $ $ $ $ $ $ $ $ $ $ $ $ $ $ $ $ $ $ $ $ $ $ $ $ $ $ $ $ $ $ $ $ $ $ $ $ $ $ $ $ $ $ $ $ $ $ $ $ $ $ $ $ $ $ $ $ $ $ $ $ $ $ $ $ $ $ $ $ $ $ $ $ $ $ $ $ $ $ $ $ $ $ $ $ $ $ $ $ $ $ $ $ $ $ $ $ $ $ $ $ $ $ $ $ $ $measurement\_model(z_t, \chi_{t}, \mathcal{M})$
		\State $\chi_{t} \leftarrow resample(\chi_{t}, w_{fast}, w_{slow})$
		\State $\mathcal{K} \leftarrow KMeansClustering(\chi_{t}, n, k)$
		\State $x_t, y_t, \theta_t$ $ $ $ $ $ \leftarrow $ $ $ $ $
		$\smash{\displaystyle\max_{weight}} \mathcal{K}$
		\State \Return $x_t, y_t, \theta_t$
		\EndFunction
		\State \textbf{end function}
	\end{algorithmic}
	\label{mcl}
\end{algorithm}

\begin{algorithm}
	\caption{Occupancy Grid Mapping}
	\begin{algorithmic}[1]
		\Function{Occupancy\_Grid\_Mapping}{$z_t, x(t), y(t),$ \vspace{1pt} $ $ $ $ $ $ $ $ $\theta(t), \mathcal{M}$}
		\For{$n \leftarrow 1$ to $N$}
		\State calculate $x_t$ and $y_t$ using eq. \ref{xlidar} and \ref{ylidar}
		\State calculate $x_t^{rotated}$ and $y_t^{rotated}$ using eq. \ref{rotatedlidar}
		\State transform $x_t^{rotated}$ and $y_t^{rotated}$ from local coordinate to global coordinate
		\State $p_x \leftarrow x(t) + x_t^{rotated}$
		\State $p_y \leftarrow y(t) + y_t^{rotated}$
		\If {$l_t^{[n]} < l_{max}$}
		\State $\mathcal{M}_{p_x,p_y} \leftarrow \mathcal{M}_{p_x,p_y} + \mathcal{L}_{occ} - \mathcal{L}_{0}$
		\State normalize $\mathcal{M}_{p_x,p_y}$ into the range of $\mathcal{L}_{min} \leq \mathcal{M}_{p_x,p_y} \leq \mathcal{L}_{max}$
		\State $scale \leftarrow 0$
		\While{$scale < (l_t^{[n]} - grid\_size)$}
		\State calculate $x_t$ and $y_t$ using eq. \ref{xlidar} and \ref{ylidar} based on ($scale - \frac{grid\_size}{2}$)
		\State calculate $x_t^{rotated}$ and $y_t^{rotated}$ using eq. \ref{rotatedlidar}
		\State transform $x_t^{rotated}$ and $y_t^{rotated}$ from local coordinate to global coordinate
		\State $p_x \leftarrow x(t) + x_t^{rotated}$
		\State $p_y \leftarrow y(t) + y_t^{rotated}$
		\State $\mathcal{M}_{p_x,p_y} \leftarrow \mathcal{M}_{p_x,p_y} + \mathcal{L}_{free} - \mathcal{L}_{0}$
		\State normalize $\mathcal{M}_{p_x,p_y}$ into the range of $\mathcal{L}_{min} \leq \mathcal{M}_{p_x,p_y} \leq \mathcal{L}_{max}$
		\State $scale \leftarrow scale + grid\_size$
		\EndWhile
		\State \textbf{end while}
		\EndIf
		\State \textbf{end if}		 
		\EndFor
		\State \textbf{end for}
		\State \Return $\mathcal{M}$
		\EndFunction
		\State \textbf{end function}
	\end{algorithmic}
	\label{mapping}
\end{algorithm}
\begin{algorithm}
	\caption{Simultaneous Localization and Mapping}
	\begin{algorithmic}[1]
		\Function{SLAM}{$v_x(t), v_y(t), \omega(t), z_t,$ \vspace{1pt} $\chi_{t-1}, \mathcal{M}$}
		\State $\chi_{t} \leftarrow motion\_model(v_x(t), v_y(t), \omega(t), \chi_{t-1})$
		\State $x(t), y(t), \theta(t) \leftarrow $ $pose(\smash{\displaystyle\max_{weight}} {\chi_{t}})$
		\State $\mathcal{M} \leftarrow$ $ $ $ $ $ $ $ $ $ $ $ $ $ $ $ $ $ $ $ $ $ $ $ $ $ $ $ $ $ $ $ $ $ $ $ $ $ $ $ $ $ $ $ $ $ $ $ $ $ $ $ $ $ $ $ $ $ $ $ $ $ $ $ $ $ $ $ $ $ $ $ $ $ $ $ $ $ $ $ $ $ $ $ $ $ $ $ $ $ $ $ $ $ $ $ $ $ $ $ $ $ $ $ $ $ $ $ $ $ $ $ $ $ $ $ $ $ $ $ $ $ $ $ $ $ $ $ $ $ $ $ $ $ $ $ $ $ $ $ $ $ $ $ $ $ $ $ $ $ $ $ $ $ $ $ $ $ $ $ $ $ $ $ $ $ occupancy\_grid\_mapping(z_t, x(t), y(t),$ \vspace{1pt} $\theta(t), \mathcal{M} )$
		\State $\chi_{t}, w_{fast}, w_{slow}$ $\leftarrow$  $measurement\_model(z_t, \chi_{t}, \mathcal{M})$
		\State $\chi_{t} \leftarrow resample(\chi_{t}, w_{fast}, w_{slow})$
		\State $\mathcal{K} \leftarrow KMeansClustering(\chi_{t}, n, k)$
		\State $x_t, y_t, \theta_t$ $ $ $ $ $ $ $ $ $ $ $ $ $ \leftarrow $ $ $ $ $ $ $ $ $ $ $
		$\smash{\displaystyle\max_{weight}} \mathcal{K}$
		\State \Return $x_t, y_t, \theta_t, \mathcal{M}$
		\EndFunction
		\State \textbf{end function}
	\end{algorithmic}
	\label{slam}
\end{algorithm}

\hspace{100pt}
\vspace{100pt}
\hspace{100pt}
\vspace{100pt}

\begin{figure}[htbp]
	\centerline{\includegraphics[width=60mm,scale=1.0]{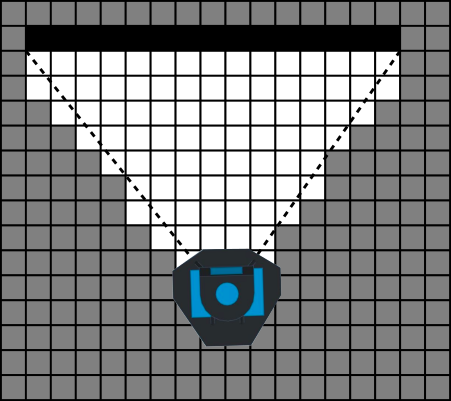}}
	\caption{Occupancy Grid Mapping: Unknown Area (gray), Obstacle Free Area (white) and Obstacle (black). }
	\label{occupancygridmap}
\end{figure}

As mentioned before, the LIDAR data $z_t$ is required in occupancy grid mapping as the core of calculation in order to construct a complete map $\mathcal{M}$. As shown in the figure \ref{occupancygridmap}, the map is in the form of the group of the grid which has the likelihood value of uncertainty that inherent in the sensor. The map is updated by using log-odds values $\mathcal{L}_{occ}, \mathcal{L}_{free}$ and $\mathcal{L}_0$. The log-odds value could be defined as the logarithm value of the ratio of the probability of a particular event and its opposite. In this case, the particular event is the event of the grid is being occupied. In other words, the log-odds value measures the likelihood that the grid is being occupied. The log-odds value could be formulated as follows.
\begin{align}
\label{log-odds}
l = \log{\frac{p(x=occupied)}{1 - p(x=occupied)}}
\end{align}

The parameter $\mathcal{L}_{occ}$ is the log-odds value on the grid which observed as occupied grid. On the other hand,  $\mathcal{L}_{free}$ is the log-odds value on the grid which observed as free grid. These values are calculated for the observed grid which is scanned by the sensor. The scanned grid also subtracted by $\mathcal{L}_0$ which is the log-odds value that regulates uncertainty of the map in regard to the sensor probabilistic characteristic. The algorithm itself also requires the position of the robot at time $t$. The position of the robot is obtained from the position of the particle which has the highest weight value. The process of constructing map is done before the $measurement\_model$ of Monte Carlo Localization. Then, the concept of the SLAM based on particle filter and occupancy grid mapping can be defined by a repeated series of $motion\_model$, constructing map, $measurement\_model$ and resampling process \cite{b20}. The occupancy grid mapping algorithm is shown in algorithm \ref{mapping} and the SLAM is shown in algorithm \ref{slam}.

\subsection{World Model}
This {\fontsize{9}{1.2}{\fontfamily{qcr}{\selectfont {node}}}} creates a complete model of the environment that has been perceived by the robot. There are two kinds of environments that are processed in this {\fontsize{9}{1.2}{\fontfamily{qcr}{\selectfont {node}}}}, namely the external environment and internal environment. The external environment is data in the form of a map which is acquired from {\fontsize{9}{1.2}{\fontfamily{qcr}{\selectfont {$/$mapinfo}}}} {\fontsize{9}{1.2}{\fontfamily{qcr}{\selectfont {topic}}}}. On the other hand, the internal environment is data in the form of the calculated pose of the robot which is acquired from {\fontsize{9}{1.2}{\fontfamily{qcr}{\selectfont {$/$robot$\_$localization}}}} {\fontsize{9}{1.2}{\fontfamily{qcr}{\selectfont {topic}}}}. The pose data is consists of position at $x$-axis in meter, position at $y$-axis in meter and rotation about $z$-axis in radian and degree. All of these data are processed and gathered intending to be published through {\fontsize{9}{1.2}{\fontfamily{qcr}{\selectfont {$/$robot$\_$worldmodel}}}} {\fontsize{9}{1.2}{\fontfamily{qcr}{\selectfont {topic}}}}. 

\subsection{Path Planning}
\label{pathplanning}
This {\fontsize{9}{1.2}{\fontfamily{qcr}{\selectfont {node}}}} generates the path from the initial point to the desired location which the robot should follow. The path is sent through {\fontsize{9}{1.2}{\fontfamily{qcr}{\selectfont {$/$robot\_pathplanning}}}} {\fontsize{9}{1.2}{\fontfamily{qcr}{\selectfont {topic}}}} which is requested by setting the desired location via {\fontsize{9}{1.2}{\fontfamily{qcr}{\selectfont {$/$goal\_service}}}} {\fontsize{9}{1.2}{\fontfamily{qcr}{\selectfont {topic}}}}. In order to generate path $\mathcal{P}$ which is not required to create the graph decomposition of the map to make the plausible connection amongst the nodes or in another word does not necessary to know the complete size of the map which is compatible with the dynamic map update in \hyperref[SLAM]{\textit{SLAM}}, the graph-based path planning algorithm which is known as Rapidly-exploring Random Tree (RRT) could be used \cite{b21}. RRT starts with building a tree usually initialize with an initial node $q_{init}$ and then adding a node to the tree connected through the edge until a certain condition, in this case, can be the number of iterations of $K$ applied to the algorithm or there is a node closest to the destination within a certain stop radius $\varDelta stop$. 

\begin{figure}[htbp]
	
	\begin{subfigure}{0.5\textwidth}
		\centerline{\includegraphics[width=0.4\linewidth, height=3cm]{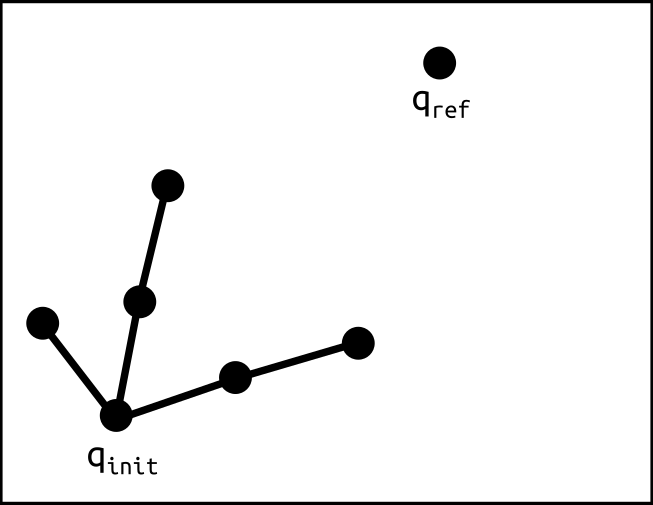}\includegraphics[width=0.4\linewidth, height=3cm]{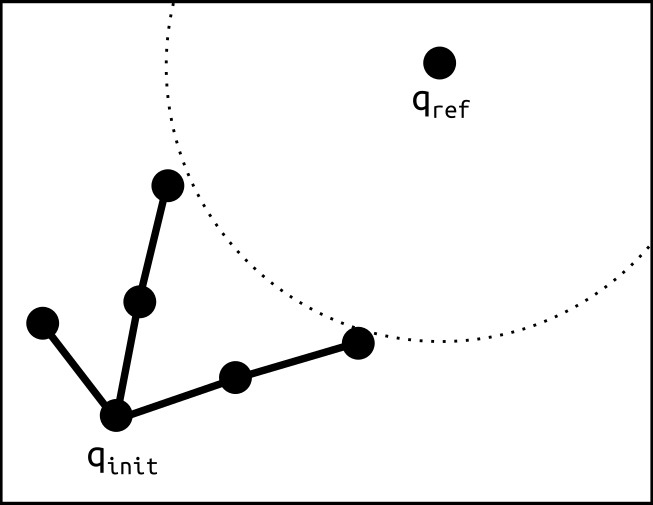}}
		\caption{$q_{ref}$ is sampled and searching for the nearest node of $q_{ref}$.}
		\label{rrt-01}
		\vspace{12pt}
	\end{subfigure}

	\begin{subfigure}{0.5\textwidth}
		\centerline{\includegraphics[width=0.4\linewidth, height=3cm]{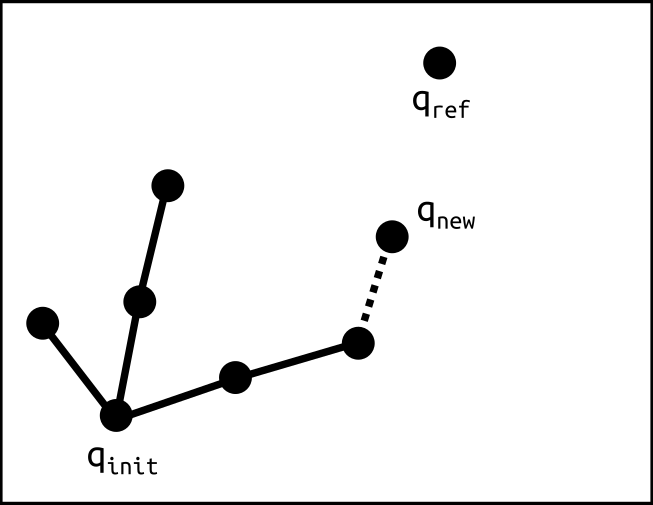}\includegraphics[width=0.4\linewidth, height=3cm]{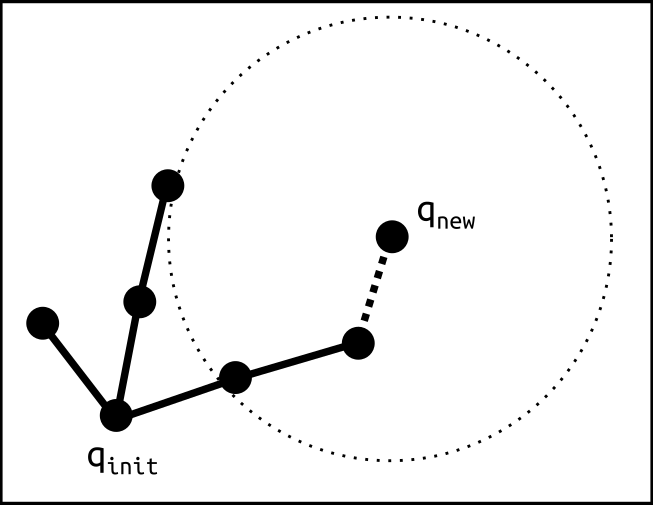}}
		\caption{$q_{new}$ is determined and searching for the nearest nodes of $q_{new}$ for the tree improvement.}
		\label{rrt-02}
		\vspace{12pt}
	\end{subfigure}

	\begin{subfigure}{0.5\textwidth}
		\centerline{\includegraphics[width=0.4\linewidth, height=3cm]{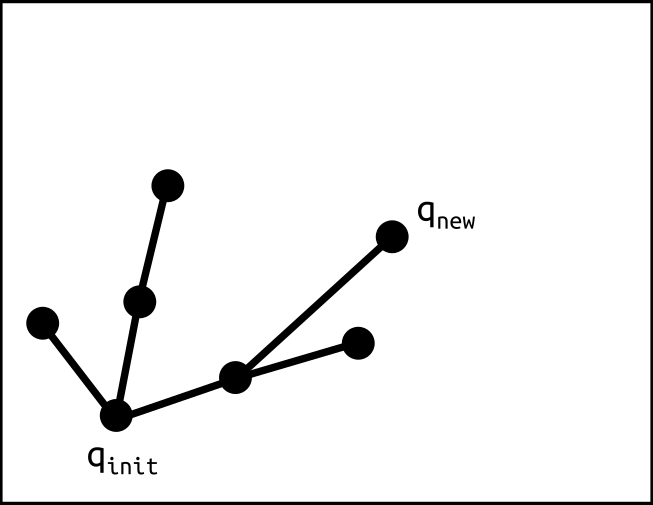}}
		\caption{$q_{new}$ is added to the tree.}
		\label{rrt-05}
	\end{subfigure}
	\caption{Growth Process of RRT*.}
	\label{rrtstar}
\end{figure}

During the iterative endeavor to grow the tree, random node $q_{ref}$ is sampled with a uniform distribution that constrained in the range of $q_{init} \pm \varphi$ to $q_{goal} \pm \varphi$. Then, the node is tested whether it is inside the map and it is not a part of the obstacles or not. The nearest node $q_{near}$ is drawn from the tree based on $q_{ref}$ that passed the test. The $q_{near}$ is used as the origin of the unit vector which references to $q_{ref}$. In order to get the new node $q_{new}$, the unit vector is expanded with the magnitude that sampled from uniform distribution in the range of stop radius $\varDelta_{stop}$ to maximum step size $\varDelta_{step}$. Between $q_{near}$ and $q_{new}$ must be free from the obstacle in the appropriate range. Then, $q_{new}$ parent is assigned with $q_{near}$. Thus, the path is obtained by backtracking the parent from the node $q_{end}$, the nearest node from the goal node $q_{goal}$, until the initial node $q_{init}$.

\begin{algorithm}
	\caption{RRT*}
	\begin{algorithmic}[1]
		\Function{RRT*}{$q_{init}, q_{goal}, K, \mathcal{T}$}
		\State $\mathcal{T}.init(q_{init})$
		\State $\mathcal{P} \leftarrow \{\}$
		\For{$k \leftarrow 1$ to $K$}
		\If{$dist(q_{goal}, \mathcal{T}.nearest(q_{goal})) \leq \varDelta_{stop}$}
		\State \textbf{break}
		\EndIf
		\State \textbf{end if}
		\State $q_{ref} \sim \mathcal{U}(q_{init} \pm \varphi, q_{goal}\pm \varphi)$
		\If{ $\neg InsideMap(q_{ref})$ $\lor$ $Obstacle(q_{ref})$}
		\State \textbf{continue}
		\EndIf
		\State \textbf{end if}
		\State $q_{near} \leftarrow \mathcal{T}.nearest(q_{ref})$
		\State $q_{new} \leftarrow \frac{q_{ref} - q_{near}}{dist(q_{ref}, q_{near})}$ $\cdot$ $ $ 
		$r \sim \mathcal{U}(\varDelta_{stop}, \varDelta_{step})$
		\If{ $\neg ObstacleFree(q_{near}, q_{new})$}
		\State \textbf{continue}
		\EndIf
		\State \textbf{end if}
		\State $\mathcal{Q} \leftarrow \{\}$
		\State $parent(q_{new}) \leftarrow q_{near}$
		\State $cost(q_{new}) \leftarrow dist(q_{new}, q_{near}) + cost(q_{near})$
		\State $\mathcal{Q} \cup \mathcal{T}.nearest(q_{new})$ in the radius of $r_{near}$
		\For{$q \in \mathcal{Q}$}
		\State $c \leftarrow cost(q)+dist(q_{new},q)$
		\If {$c < cost(q_{new})$}
		\If {$ObstacleFree(q, q_{new})$}
		\State $cost(q_{new}) \leftarrow c$
		\State $parent(q_{new}) \leftarrow q$
		\EndIf
		\State \textbf{end if}
		\EndIf
		\State \textbf{end if}
		\EndFor
		\State \textbf{end for}
		\State $\mathcal{T}.grow(q_{new})$ 
		\EndFor
		\State \textbf{end for}
		\State $q_{end} \leftarrow \mathcal{T}.nearest(q_{goal})$
		\While {$parent(q_{end})$}
		\State $\mathcal{P} \cup q_{end}$
		\State $q_{end} \leftarrow parent(q_{end})$
		\EndWhile
		\State \textbf{end while}
		\State \Return $\mathcal{P}$
		\EndFunction
		\State \textbf{end function}
	\end{algorithmic}
	\label{rrt}
\end{algorithm}

RRT can be improved in a heuristic way to update the parent of $q_{new}$ in order to get a shorter and smoother path. The concept of this method is to find a node in a set $\mathcal{Q}$ consists of several nearest nodes $q$ in a certain radius $r_{near}$ which has the cost combined with the distance to $q_{new}$ better than the cost of $q_{new}$. This node is the new parent for the $q_{new}$. This improved RRT is called RRT* \cite{b22}. The RRT* algorithm is shown in algorithm \ref{rrt} and visualized in figure \ref{rrtstar}. The tree $\mathcal{T}$ which is one of the essential components in RRT could be stored in the form of $k$-d tree \cite{b23}. It is the developed binary search tree to store multidimensional data such as the robot position in $x_t$ and $y_t$. This tree is used because of its running time of searching the nearest node is faster, $O(\log n)$, than linear search, $O(n)$, which is required in real-time path planning\cite{b24}.

\subsection{Robot Control}
This {\fontsize{9}{1.2}{\fontfamily{qcr}{\selectfont {node}}}} attempts to drive the robot to the desired setpoint. This desired point is a pose $(x, y, \theta)$ obtained from the path via {\fontsize{9}{1.2}{\fontfamily{qcr}{\selectfont {$/$robot\_pathplanning}}}} {\fontsize{9}{1.2}{\fontfamily{qcr}{\selectfont {topic}}}}. The path is given when this {\fontsize{9}{1.2}{\fontfamily{qcr}{\selectfont {node}}}} request a certain location as the goal location through {\fontsize{9}{1.2}{\fontfamily{qcr}{\selectfont {$/$goal\_service}}}} {\fontsize{9}{1.2}{\fontfamily{qcr}{\selectfont {topic}}}}. This received data in regard to the desired location is determined by \hyperref[behaviour]{\textit{Behaviour}} via {\fontsize{9}{1.2}{\fontfamily{qcr}{\selectfont {$/$goal\_service\_control}}}} {\fontsize{9}{1.2}{\fontfamily{qcr}{\selectfont {topic}}}}. Then, this {\fontsize{9}{1.2}{\fontfamily{qcr}{\selectfont {node}}}} keep updating the path after certain steps in order to overcome the dynamic environment problem. The path also updated whenever received blocked flag from {\fontsize{9}{1.2}{\fontfamily{qcr}{\selectfont {$/$blocked\_path\_service}}}} {\fontsize{9}{1.2}{\fontfamily{qcr}{\selectfont {topic}}}}. 

The robot is controlled to achieve desired point by the most widely used control design technique, PID controller \cite{b25}. In this {\fontsize{9}{1.2}{\fontfamily{qcr}{\selectfont {node}}}}, PD controller, the variety of PID controller, is used by tuning the proportional term $K_p$ and derivative term $K_d$ to appropriate value in an effort to gain the robot stability. The PD controller proved to have a decent performance for mobile robot position control with its nature on decreasing the settling time and overshoot  \cite{b26}. This kind of control system method could alleviate the complexity of the tuning process because solely concern with two parameters. Thus, the PD controller can be formulated as follows.
\begin{align}
u(t) = K_p \cdot e(t) + K_d \cdot \frac{de(t)}{dt}
\end{align}

The proportional term is emphasizing the error $e(t)$ to raise the control signal $u(t)$ proportionally. The derivative term is amplifying the rate change of the error $\frac{de(t)}{dt}$ which indicates the future value of the error would go. The error $e(t)$ is the difference between the desired point and the robot pose which is obtained from {\fontsize{9}{1.2}{\fontfamily{qcr}{\selectfont {$/$robot\_worldmodel}}}} {\fontsize{9}{1.2}{\fontfamily{qcr}{\selectfont {topic}}}} then compute the feedback loop from it. The rate change of error $\frac{de(t)}{dt}$ could be defined as the difference between $e(t)$ and $e(t-1)$ in the interval time  $t$. The control signals $u(t)$ in the form of $v_x(t), v_y(t)$ and $\omega(t)$ are published through \fontsize{9}{1.2}{\fontfamily{qcr}{\selectfont {/cmd\_vel}}} {\fontsize{9}{1.2}{\fontfamily{qcr}{\selectfont {topic}}}}.

\subsection{Communication}
This {\fontsize{9}{1.2}{\fontfamily{qcr}{\selectfont {node}}}} handles communication system which regulates the robot internal data. The data, consists of the robot pose data and the command to the desired room from HRI, is stored in a certain database system. By storing the data into the database, the data could be monitored remotely in order to know the most up-to-date robot's internal data. The command from HRI is processed to be sent as a request through {\fontsize{9}{1.2}{\fontfamily{qcr}{\selectfont {$/$goal\_service\_behaviour}}}} {\fontsize{9}{1.2}{\fontfamily{qcr}{\selectfont {topic}}}}. This {\fontsize{9}{1.2}{\fontfamily{qcr}{\selectfont {node}}}} has the robot pose data which subscribed from {\fontsize{9}{1.2}{\fontfamily{qcr}{\selectfont {$/$robot\_worldmodel}}}} {\fontsize{9}{1.2}{\fontfamily{qcr}{\selectfont {topic}}}}.

The communication system makes use of the Firebase Realtime Database, a cloud database system based on the NoSQL database which the client can update the data inside the database through HTTP request as JavaScript Object Notation (JSON) format\cite{b27}. Using the key-value pair fashion, the NoSQL database could perform Create, Read, Update, Delete (CRUD) operations faster than the relational one \cite{b28}. 

The robot pose and the command data are stored with {\fontsize{9}{1.2}{\fontfamily{qcr}{\selectfont {position}}}} and {\fontsize{9}{1.2}{\fontfamily{qcr}{\selectfont {desired\_room}}}} key inside the database respectively. The robot pose is updated whenever this {\fontsize{9}{1.2}{\fontfamily{qcr}{\selectfont {node}}}} receives the information from {\fontsize{9}{1.2}{\fontfamily{qcr}{\selectfont {$/$robot\_worldmodel}}}} {\fontsize{9}{1.2}{\fontfamily{qcr}{\selectfont {topic}}}}. Subsequentially, this {\fontsize{9}{1.2}{\fontfamily{qcr}{\selectfont {node}}}} get the updated command data from the HRI by requesting the database via {\fontsize{9}{1.2}{\fontfamily{qcr}{\selectfont {desired\_room}}}} key.

\subsection{Behaviour}
\label{behaviour}
This {\fontsize{9}{1.2}{\fontfamily{qcr}{\selectfont {node}}}} makes use of a finite-state machine (FSM) concept as the way the robot should behave which translates the given command or certain circumstances as an input of state-transition. Using this paradigm the code for this {\fontsize{9}{1.2}{\fontfamily{qcr}{\selectfont {node}}}} could be designed into a maintainable code due to its readable and understandable nature \cite{b29}. The robot has two states: {\fontsize{9}{1.2}{\fontfamily{qcr}{\selectfont {Escorting}}}} and {\fontsize{9}{1.2}{\fontfamily{qcr}{\selectfont {Homing}}}}. The {\fontsize{9}{1.2}{\fontfamily{qcr}{\selectfont {Escorting}}}} state asks the robot to escort to the desired point {\fontsize{9}{1.2}{\fontfamily{qcr}{\selectfont {goal\_pos}}} whenever the {\fontsize{9}{1.2}{\fontfamily{qcr}{\selectfont {goal\_pos}}}} is requested from {\fontsize{9}{1.2}{\fontfamily{qcr}{\selectfont {/goal\_service\_behaviour}}}} {\fontsize{9}{1.2}{\fontfamily{qcr}{\selectfont {topic}}}}. This {\fontsize{9}{1.2}{\fontfamily{qcr}{\selectfont {goal\_pos}}}} information is passed on to {\fontsize{9}{1.2}{\fontfamily{qcr}{\selectfont {/goal\_service\_control}}}} {\fontsize{9}{1.2}{\fontfamily{qcr}{\selectfont {topic}}}} as a setpoint for the robot. Then, the {\fontsize{9}{1.2}{\fontfamily{qcr}{\selectfont {Homing}}}} state tells the robot to go back to its initial point {\fontsize{9}{1.2}{\fontfamily{qcr}{\selectfont {init\_pos}}}} shortly after the robot has arrived at \fontsize{9}{1.2}{\fontfamily{qcr}{\selectfont {goal\_pos}}}}. This {\fontsize{9}{1.2}{\fontfamily{qcr}{\selectfont {node}}}} calculates the error between the current robot position and the destination point to check whether the robot has arrived at \fontsize{9}{1.2}{\fontfamily{qcr}{\selectfont {goal\_pos}}} or not by monitoring the robot position from \fontsize{9}{1.2}{\fontfamily{qcr}{\selectfont {/robot\_worldmodel}}} {\fontsize{9}{1.2}{\fontfamily{qcr}{\selectfont {topic}}}}. The state-transition table for the robot is shown in table \ref{state}.

\begin{table}[h!]
	
	\begin{center}
		\caption{State-transition Table for the Robot}
		\label{state}
		\begin{tabular}{|c|c|c|c|} 
			\hline
			\textbf{Current State} & \textbf{Input} & \textbf{Next State} & \textbf{Output}\\
			\hline
			& & & \\
		{\fontsize{8}{1.2}{\fontfamily{qcr}{\selectfont {Escorting}}}}	& arrived at  & {\fontsize{8}{1.2}{\fontfamily{qcr}{\selectfont {Homing}}}} & move to\\
		& {\fontsize{8}{1.2}{\fontfamily{qcr}{\selectfont {goal\_pos}}}} & & {\fontsize{8}{1.2}{\fontfamily{qcr}{\selectfont {init\_pos}}}}\\
		& & & \\
		\hline
		& & & \\
		{\fontsize{8}{1.2}{\fontfamily{qcr}{\selectfont {Homing}}}}	& {\fontsize{8}{1.2}{\fontfamily{qcr}{\selectfont {goal\_pos}}}} & {\fontsize{8}{1.2}{\fontfamily{qcr}{\selectfont {Escorting}}}} & move to  \\
		& is requested & & {\fontsize{8}{1.2}{\fontfamily{qcr}{\selectfont {goal\_pos}}}}\\
		& & & \\
		\hline
		
		\end{tabular}
	\end{center}
\end{table}

\section{Simulation and Experiments}
\label{SE}
\begin{figure}[htbp]
	
	\begin{subfigure}{0.5\textwidth}
		\centerline{\includegraphics[width=0.5\linewidth, height=3cm]{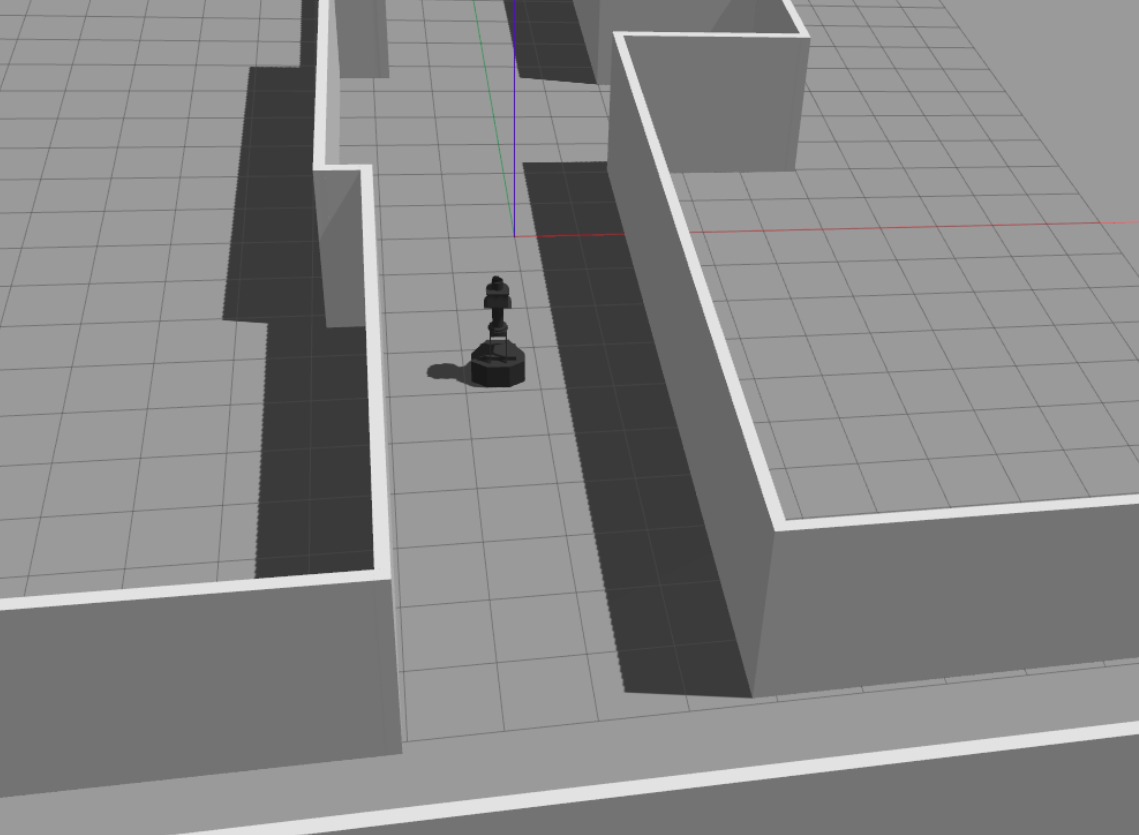}\includegraphics[width=0.5\linewidth, height=3cm]{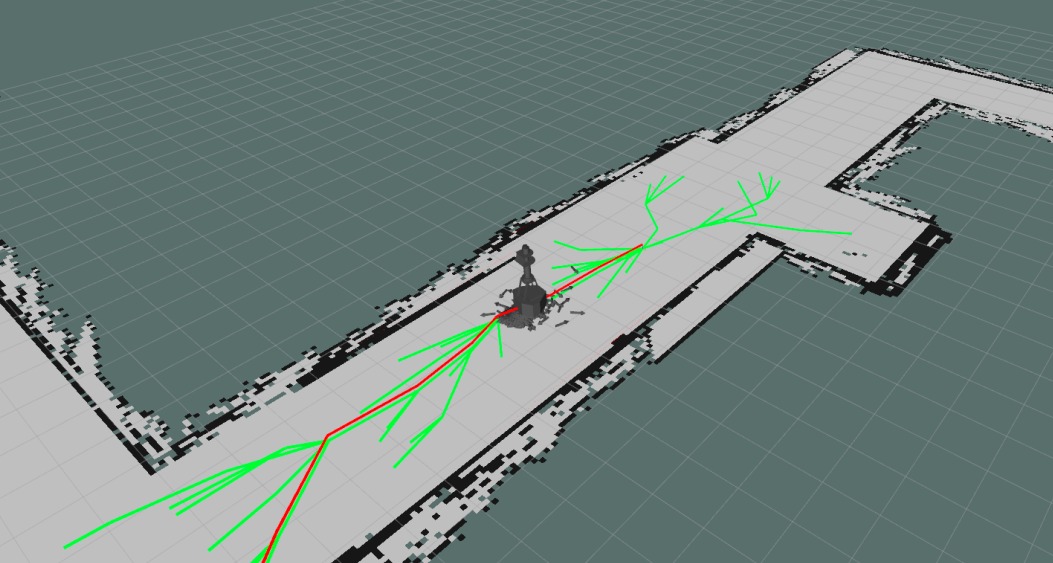}}
		\label{gazebo}
	\end{subfigure}
	
	\caption{Robot Simulation: Gazebo (left) and RViz (right).}
	\label{simulation}
\end{figure}

NightOwl simulates the robot and its environment in Gazebo as the physical realization and RViz as a visualization of the perceived and calculated data. These data are from \hyperref[SLAM]{\textit{SLAM}} which is consists of the map and distribution of particle and \hyperref[pathplanning]{\textit{Path Planning}} which is consists of the tree and path. The robot in the simulation responses the requested data 
through a certain {\fontsize{9}{1.2}{\fontfamily{qcr}{\selectfont {topic}}}} handled by the gazebo plugin which is inspired by Yao et al. \cite{b30}. The plugin is designated for the omnidirectional robot which could move the robot through {\fontsize{9}{1.2}{\fontfamily{qcr}{\selectfont {/cmd\_vel}}}} {\fontsize{9}{1.2}{\fontfamily{qcr}{\selectfont {topic}}}} and get the odometry data through {\fontsize{9}{1.2}{\fontfamily{qcr}{\selectfont {/odom}}}} {\fontsize{9}{1.2}{\fontfamily{qcr}{\selectfont {topic}}}}. The robot simulation is shown in figure \ref{simulation}.

Several experiments were conducted to test the performance of the platform on the simulation and the real hardware. The laptop PC with Intel Core i5 10\textsuperscript{th} Generation 1.6 GHz processor, 8 GB memory and Ubuntu 18.04 Operating System was used as the main controller and the Teensy 4.0 with ARM Cortex-M7 600 MHz processor, 2048 kB flash memory, 1024 kB RAM and NightOwl custom firmware as the microcontroller for the experiments. These experiments focused on sundry problems in robot navigation.

\subsection{Comparison of Generated Path}
\hyperref[pathplanning]{\textit{Path Planning}} output is a generated path from the RRT algorithm. The algorithm itself depends on a tree which is consists of many nodes in a possible position on the map. These nodes are all possible candidates for the path that the robot could follow. The tree itself could be represented as a linear array of nodes or in the form of a $k$-d tree data structure. As known before, the $k$-d presents a fast searching for the nearest neighbor of the arbitrary node. With the same destination point and maximum iteration $K = 625$, the time required for generating path is compared for the RRT with $k$-d tree and with a standard array. The result is shown in table \ref{time}.

\begin{table}[h!]
	
	\begin{center}
		\caption{Time Required for Generating Path}
		\label{time}
		\begin{tabular}{c | c | c | c}
			& Best (ms) & Average (ms) & Worst (ms) \\
			\hline
			RRT with $k$-d tree & \textbf{10.322} & \textbf{16.657} & \textbf{25.25}\\
			RRT with a standard array & $10.439$ & $17.499$ & $28.562$ \\
		\end{tabular}
	\end{center}
\end{table}

During the experiments, the paths are generated overtime after some steps. Table \ref{time} shows the best, average, and worst time for generating a path. It shows the RRT with $k$-d tree achieves the fastest time in all of the statistical performance due to its time complexity which has better performance $O(\log n)$ than the RRT with a standard array $O(n)$ when searching the nearest neighbor during the planning process.
  

\begin{figure}[htbp]
	\centerline{\includegraphics[width=80mm,scale=1.0]{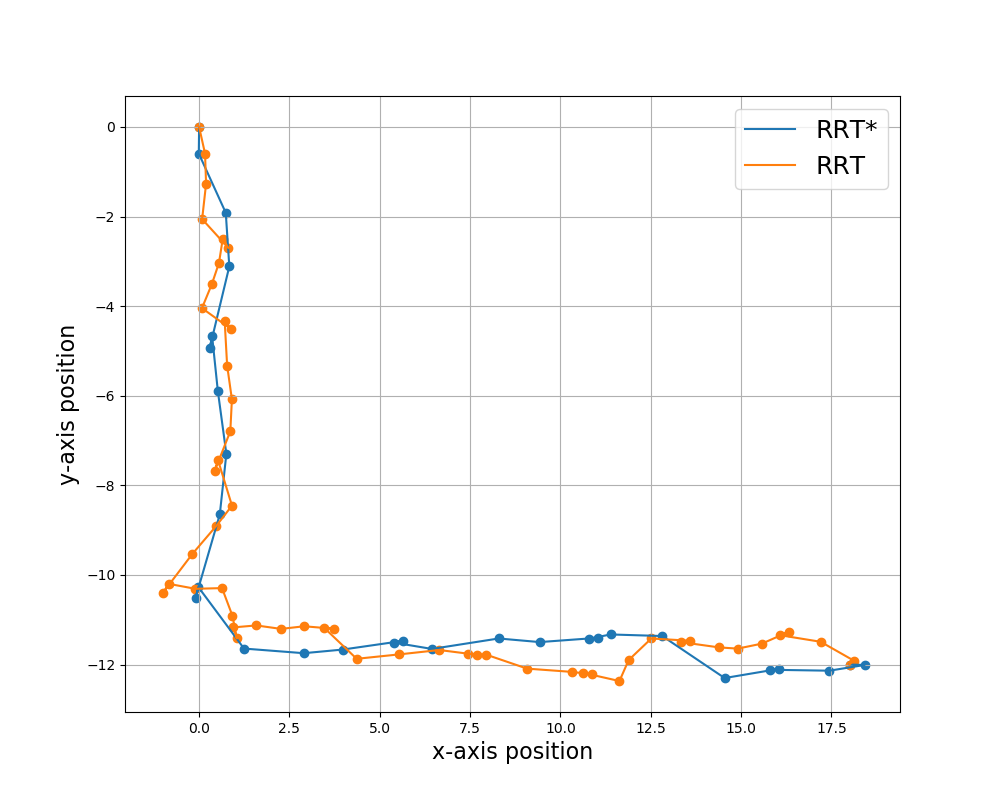}}
	\caption{Generated Path Comparison Between RRT* and RRT.}
	\label{rrt-comparison}
\end{figure}

The generated path of RRT* which is the heuristical improvement of RRT also evaluated. The RRT* attempts to make the path from the RRT shorter and smoother. The experiments were conducted within the Department of Electrical and Information Engineering of Universitas Gadjah Mada simulation environment by requesting the robot to move from the lobby to the teleconference room with no moving obstacles in the empty hallway. It is shown in figure \ref{rrt-comparison} that RRT* has a less jagged path than RRT. The jagged path makes the robot move in a dramatically unsteady way which can result in the odometry error increase drastically.

\subsection{Odometry with Path Planning Consideration}
In order to convince the prior statement, the odometry tests were carried out on the actual robot. The robot was assigned to navigate from the initial point to the desired point to observe the impact of the distinct path planning method. The robot moved from the lobby to the thesis defence room back and forth. Then, the final position was compared between the actual measurement and the odometry in $xy$-plane.

\begin{table}[h!]
	
	\begin{center}
		\caption{Odometry Results from Different Path Planning Methods}
		\label{path}
		\begin{tabular}{|c | c | c | c | c | c | c |}\hline
			Initial - & Path & \multicolumn{4}{c|}{Final Position (cm)} &  Dist.\\\cline{3-6}
			 Desired & Planning  & \multicolumn{2}{c|}{Actual} & \multicolumn{2}{c|}{Odometry} &  Error\\\cline{3-6}
			 Point & Method & $x$ & $y$ & $x$ & $y$ & (cm)\\
			\hline
			Lobby - Thesis & RRT & $0.0$ & $720.0$ & $12.0$ & $696.0$ & $26.833$ \\
			Defence Room & RRT* & $0.0$ & $720.0$ & $16.0$ & $701.0$ & $24.839$ \\\hline
			Thesis Defence & RRT & $60.0$ & $0.0$ & $0.0$ & $13.0$ & $61.392$ \\
			Room - Lobby & RRT* & $30.0$ & $0.0$ & $4.0$ & $8.0$ & $27.203$ \\\hline
		\end{tabular}
	\end{center}
\end{table}

Table \ref{path} shows the comparison of the effect between two different path planning methods on odometry at each certain scenario. The distance error is measured by applying Euclidean distance between the actual position and the odometry. It shows the distance error of the odometry with RRT is larger than the RRT* ones at each of the scenarios. These  differences imply the jagged path resulted from the RRT affects the odometry calculation so that it tends to err quickly based on hypothesis testing using t-test with significance level ($\alpha$) = $0.3$. Whilst the RRT* attempts to contrive an efficient path so that the redundant movement could be circumvented. The redundant movement is one of the facets that hasten the odometry error. The odometry error occurred by virtue of several issues for instance inaccuracy measurement of the sensors, slippage due to the sleek floor surface, and information losses during the computation.

\subsection{SLAM Result}

The \hyperref[SLAM]{\textit{SLAM}} algorithm was initially configured with the following setup:
\begin{itemize}
	\item Number of particles ($M$): $625$.
	\item All of the particles' states $(x_t, y_t, \theta_t)$ were assigned to $(0,0,0)$ which is the location of the lobby of Department of Electrical and Information Engineering.
	\item Weight of each particles: $\frac{1}{M}$
	\item Particle drift factor ($\gamma$): $0.1$
	\item The resampling parameters $(w_{slow}, w_{fast}, a_{slow}, a_{fast}, \varOmega)$ were assigned to $(0.0, 0.0, 0.0125, 62.5, 0.025)$ respectively.
	\item Error offset of the particle resampling orientation $(\theta_\epsilon)$: \ang{1.414}.
	\item The $k$-means clustering parameters: the number of epochs and centroids $(n$ and $k)$ were assigned to $(100$ and $3)$ respectively.
	\item The inherent uncertainty characteristic of the average
	velocities of the robot $(\bar{\sigma_x}, \bar{\sigma_y}, \bar{\sigma_\theta})$ were assigned to $(0.25, 0.25, 0.0)$ respectively. 
	\item The uncertainty of the velocities that might be reached $(\sigma_{x}, \sigma_{y}, \sigma_{\theta})$ were assigned to $(1.5, 1.5, 0.005)$ respectively.
	\item The Size of each grid in the map: $10cm \times 10cm$.
	\item The log-odds value of occupancy grid mapping $(\mathcal{L}_{free}, \mathcal{L}_{occ}, \mathcal{L}_{0})$ were assigned to $(-0.5,0.5,0.001)$ and $(-1.94591,0.57536,0.00492)$ as the independent variables for the quality testing of the resulted map.
\end{itemize}

\begin{figure}[htbp]
	\begin{subfigure}{0.5\textwidth}
		\centerline{\includegraphics[width=0.6\linewidth, height=5cm]{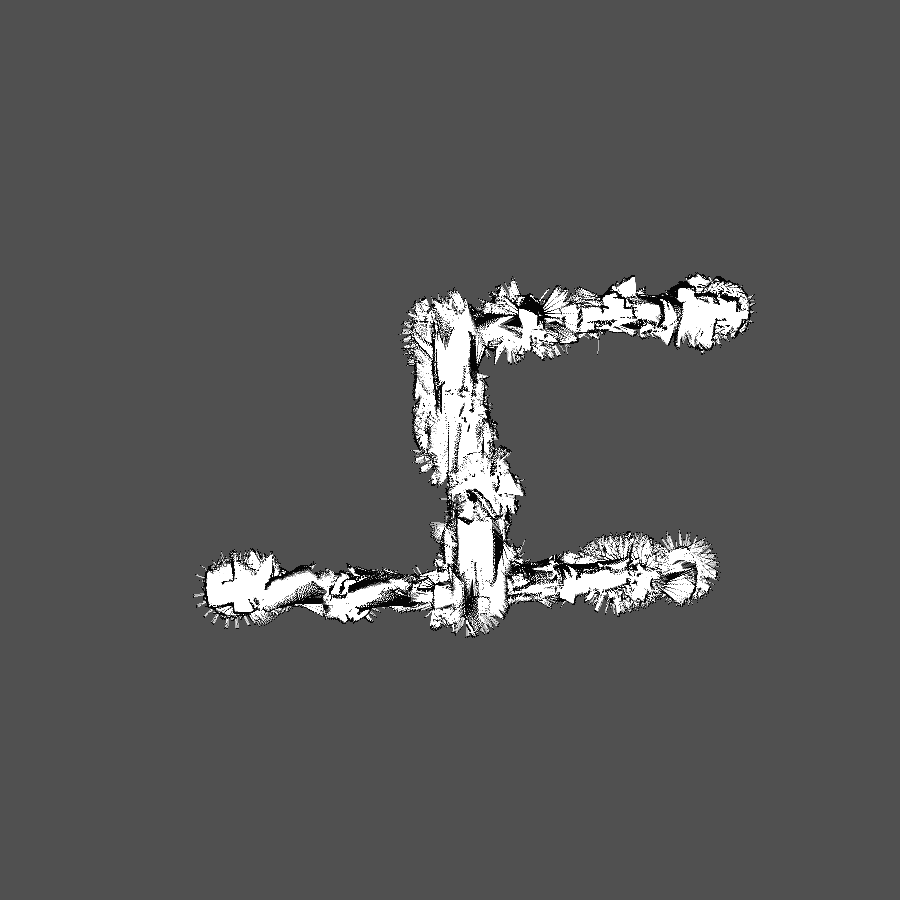}}
		\caption{$\mathcal{L}_{free} = -0.5, \mathcal{L}_{occ} = 0.5, $ and $\mathcal{L}_{0} = 0.001$.\\ }
		\vspace{10pt}
		\label{map2}
	\end{subfigure}
    
	\begin{subfigure}{0.5\textwidth}
		\centerline{\includegraphics[width=0.6\linewidth, height=5cm]{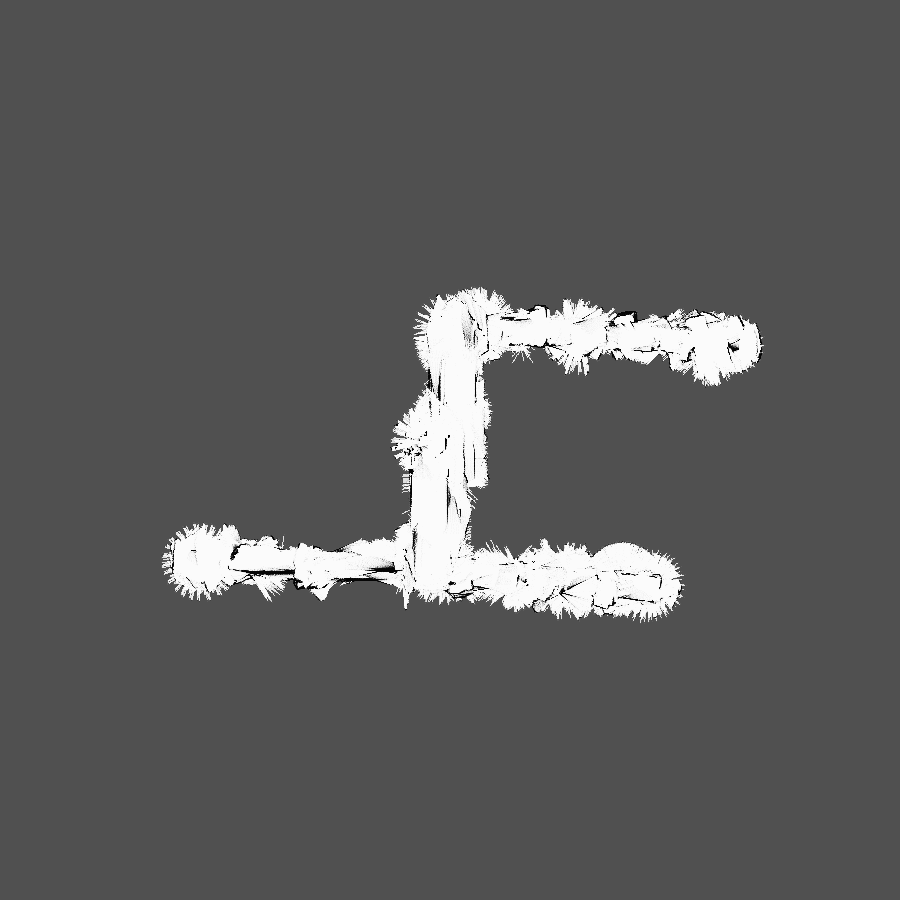}}
		\caption{$\mathcal{L}_{free} = -1.94591, \mathcal{L}_{occ} = 0.57536,$ and $\mathcal{L}_{0} = 0.00492$.}
		\label{map3}
	\end{subfigure}
	\caption{Constructed Maps of the 1\textsuperscript{st} Floor of Department of Electrical and Information Engineering.}
	\label{constructed-map}
\end{figure}

\begin{figure}[htbp]
	\centerline{\includegraphics[width=60mm,scale=1.0]{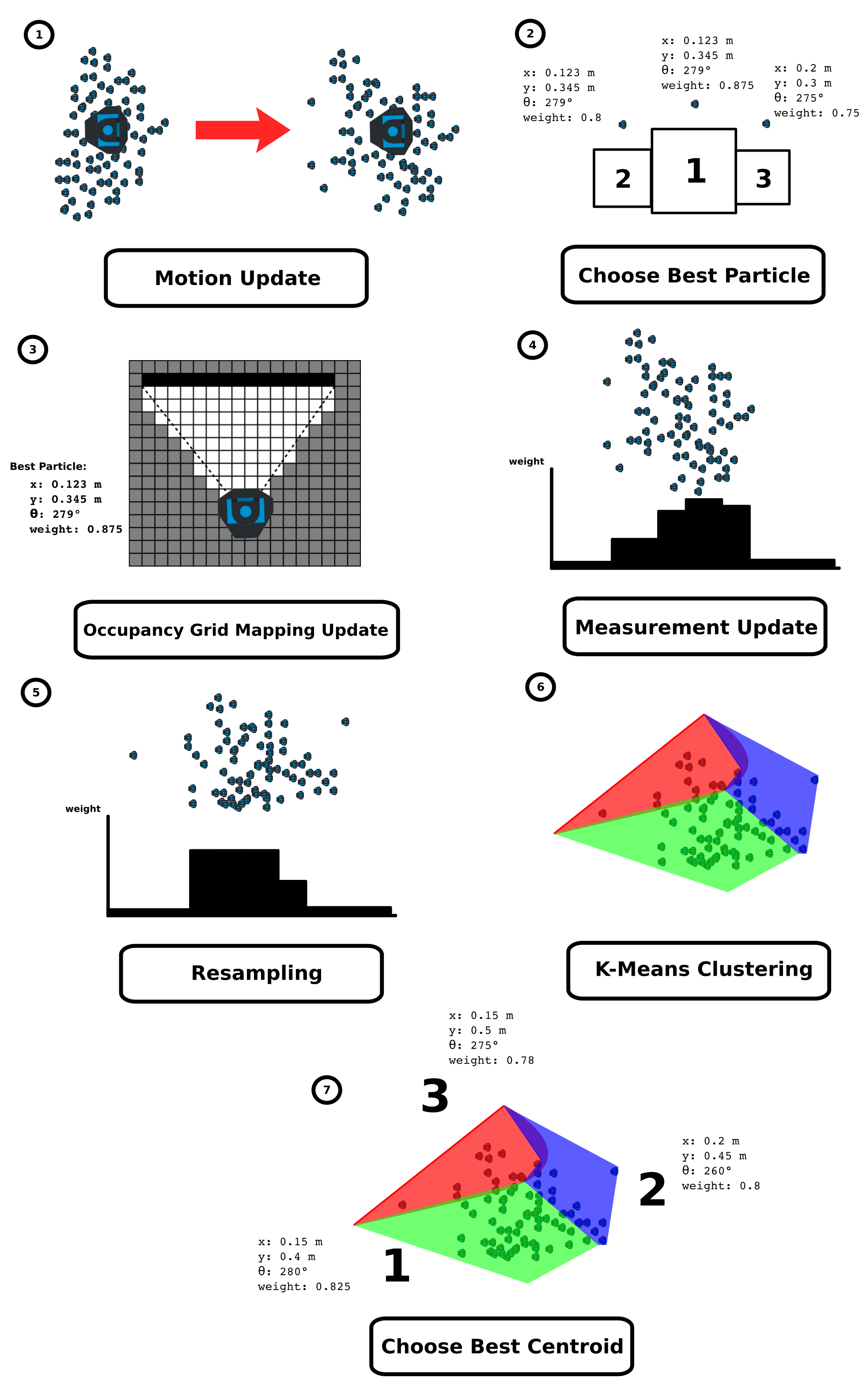}}
	\caption{SLAM Process.}
	\label{slam-process}
\end{figure}

\hyperref[SLAM]{\textit{SLAM}} was built by modifying the tinySLAM\cite{b20} and produces a map and estimated pose at the map accordingly which the process is shown in figure \ref{slam-process}. The experiments were performed using the real robot on the first floor of the  Department of Electrical and Information Engineering UGM. The SLAMTEC RPLIDAR A1 \ang{360} with a detection range from 0.15 to 12 meters was used in the experiments. The LIDAR solely perceives the objects in two-dimensional $xy$-plane at the same height which means everything below and above the LIDAR is undetected. The LIDAR reading was very noisy due to many unexpected objects jumbled in the vicinity of the hallway. These objects were excluded in the simulation which made the actual environment more challenging.

The constructed maps in figure \ref{constructed-map} show the result of the experiments. The white regions are the explored areas, the black regions are the obstacles and the gray regions are the unknown areas. Most of the parts of the environment are not surrounded by thick walls rather the object that could refract the LIDAR ray such as glass doors or windows.
Using different settings for the log-odds values, the constructed maps exhibit the contrast results. The noise due to the aforementioned circumstances gives uncertainty to determine the existent of the obstacles. Thus, the very bottom of the configuration of the log-odds values with the tendency of the obstacles are non-existent produces the best result.

\section{Conclusions and Future Works}
\label{C}
This robotic platform for a wheeled service robot, NightOwl, was introduced and elaborated in this paper. The platform utilizes the asynchronous parallel information exchange provided by ROS so that each of the {\fontsize{9}{1.2}{\fontfamily{qcr}{\selectfont {nodes}}}} could run concurrently. The robot's main quest is to take the guests who demand assistance to the coveted location. Thus, the robot navigation capability provided in the platform is necessary to accomplish the objective. The platform also provides simulation to visualize the underlying algorithm and replicate the possible situation in the actual environment.

The robot uses the RRT* path planning method which can reduce the serious impact of the accumulated increase in odometric errors compared to the RRT path planning method. One aspect that can increase the odometric errors is the redundant movement which can be resulted from the jagged path. The RRT* path planning method combined with the $k$-d tree as a tree representation accelerates the process of generating paths compared to standard array data structures. It is caused by the time complexity of the k-d tree $O(\log n)$ is more efficient than the standard array $O(n)$. Then, the quality of the maps produced by \hyperref[SLAM]{\textit{SLAM}} is affected by the log-odds value. The more the log-odds value tends to the absence of obstacles around the robot, the better the quality of the resulted map.

There are several problems in navigation that appear during this works such as the deprivation of reliable features for the \hyperref[SLAM]{\textit{SLAM}} over time and the collision due to the copiousness of the unperceived obstacles. The features deprivation situation is caused by sundry circumstances such as moving objects, translucent entities, and any other unwanted things. For the next stage development, the robot will be equipped with the camera as the sensor alongside LIDAR, IMU, rotary encoders, and IR proximity sensor. The camera will enhance the \hyperref[SLAM]{\textit{SLAM}} by capturing more reliable features from the existing vicinity objects periodically. The camera also could help to detect the obstacles which caused the collision. These obstacles located below the LIDAR position and the IR proximity sensor fails to detect its presence.

For future works, the simulation also needs to be improved. In order to simulate the noise in the actual environment, the notable obstacles such as the moving objects and the translucent things that should be replicated to anticipate the unexpected behavior of the sensors. The actual environment might not as elegant as the simulation yet it is usually cluttered and unpredictable with the existence of various entities and its inherent particular characteristics.


\end{document}